# A Formal Description of Sorani Kurdish Morphology

Sina Ahmadi
Insight Centre for Data Analytics
National University of Ireland Galway
ahmadi.sina@outlook.com

## Contents

**Abstract**

Sorani Kurdish, also known as Central Kurdish, has a complex morphology, particularly due to the patterns in which morphemes appear. Although several aspects of Kurdish morphology have been studied, such as pronominal endoclitics and Izafa constructions, Sorani Kurdish morphology has received trivial attention in computational linguistics. Moreover, some morphemes, such as the emphasis endoclitic *=îş*, and derivational morphemes have not been previously studied. To tackle the complex morphology of Sorani, we provide a thorough description of Sorani Kurdish morphological and morphophonological constructions in a formal way such that they can be used as finite-state transducers for morphological analysis and synthesis.

It should be noted that the current manuscript is being completed. Any suggestions or reporting of errors will be highly appreciated.

***Keywords***— Sorani Kurdish, Central Kurdish, formal morphology, inflectional morphology, lexeme-based morphology, computational linguistics

# 1 Introduction

In linguistics, morphology is the science of word formation. In a dictionary, words are described in their unique canonical form, that is *lemma*. However, a word can appear in several other forms based on its role in a sentence by being accompanied by other *morphemes*. A morpheme is a unit of a language that cannot be further divided, for instance that is "*-ing*" in the English word *playing*. Understanding the mechanism of creating new word forms requires a thorough knowledge of the language which plays a fundamental role in various linguistic fields, particularly syntax. Therefore, as a starting point in creating natural language processing (NLP) and computational linguistics (CL) tools and resources, language morphological analysis is being addressed as one of the first steps. Unlike richly-resourced languages, such as English and French, for which various aspects of morphology and syntax have been analyzed, many less-resourced languages such as Kurdish have received little attention in this realm.

In this paper, we present the morphology of Sorani Kurdish, one of the main variants of Kurdish, from a computational perspective. By computational, we refer to a non-verbal description where morphological processes are defined using formal rules. The main advantage of this approach is to provide a reference to enable future researchers to analyze the Kurdish morphology, and to some extent the syntax, for language technology tasks. Such a description of morphology as a key component in NLP will pave the way for further developments such as morphological analysis and spell checking.

In this paper, we first provide a general description of the Kurdish language in Section 2 and some of the major previous studies in Kurdish morphology in Section 3. In Section 4, we provide the preliminary definitions which are used throughout the manuscript. The major contributions of the paper come in Section 5, where the morphology of Sorani Kurdish is described, and in Sections 6 and 7, where morphophonological and morphological rules for analyzing Sorani Kurdish are provided using a morpheme-based approach. These rules can be considered finite-state transducers and can be used for various applications, especially spelling error correction and morphological analysis.

It is worth noting that Sorani Kurdish is spoken in many regions in diverse sub-dialects. Therefore, there are many morphological variations that should be studied separately. Our focus in this paper is on the core morphological characteristics of Sorani Kurdish which are common among sub-dialects and widely accepted in the language form used on daily basis in the media and the press. We hope that this work paves the way for more detailed documentation of morphological and syntactic variations among such sub-dialects, such as Ardalani, Babani and Mukriani.

## 2 Sorani Kurdish

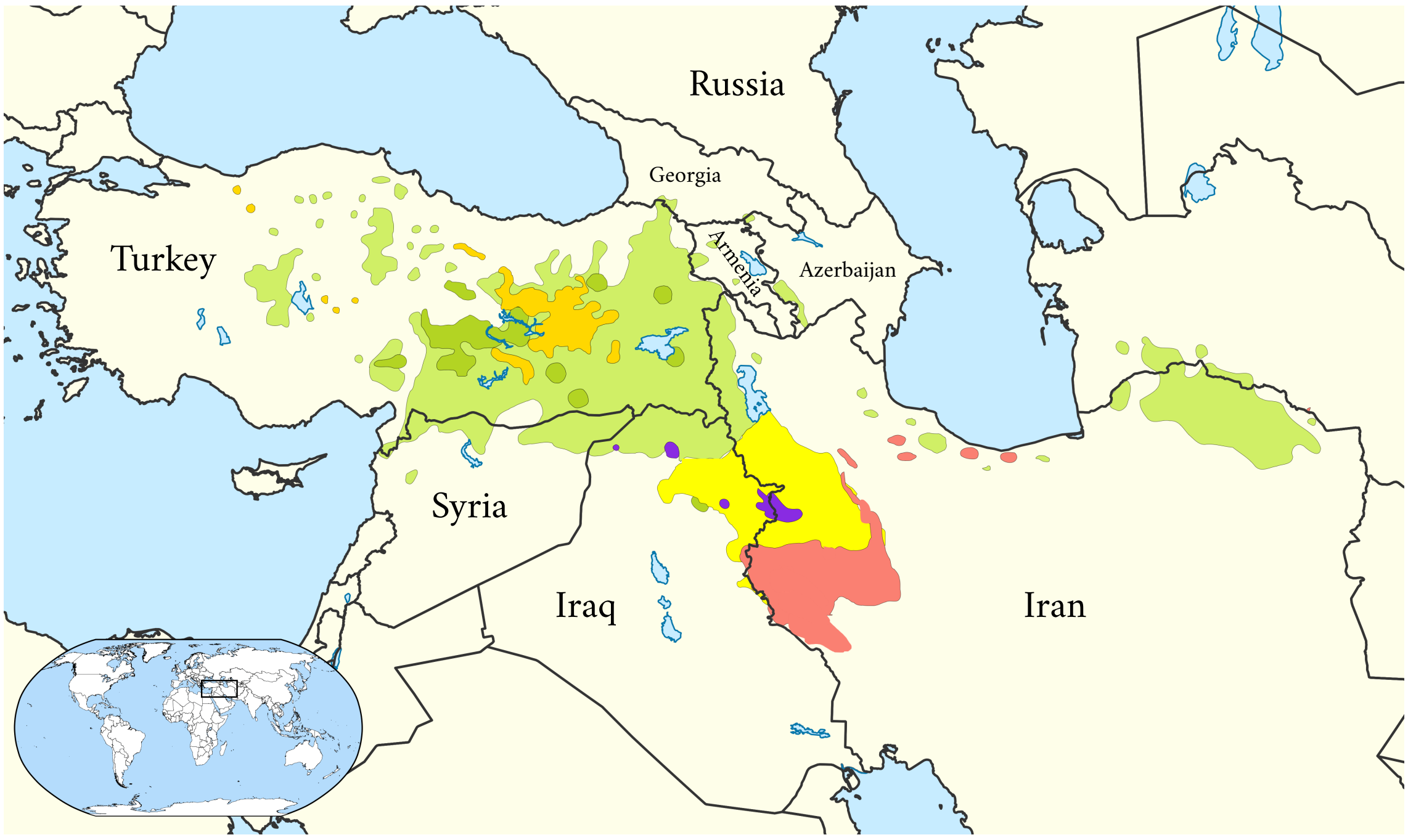

Figure 1: Geographical distribution of the dialects and languages in Kurdish regions as Northern ▢, Central ▢, Southern ▢, Zaza ▢, Gorani ▢ and Mixed areas ▢ Our focus in this manuscript is on the dialect spoken in the yellow region. (Created based on [Rekacewicz, 2008])

.

### 2.1 General Presentation

Kurdish is an Indo-European language spoken by over 25 million speakers in the Kurdish regions in Turkey, Iran, Iraq and Syria, and also by the Kurdish diaspora around the world [McCarus, 2007]. The classification of the Kurdish dialects has been a matter of debate, which mainly focuses on how to distinguish dialects as Kurdish dialects and languages as Kurdish languages. Although the standard form of the Kurdish language has been a matter of discussion within the academic communities [MacKenzie, 1962], there is no consensus upon what is meant by the Kurdish standard language. Figure 1 illustrates the geographical distributions of Kurdish dialects in the Middle East.

Central Kurdish, also known by its endonym Sorani, is mostly spoken by the Kurds within the Iranian and Iraqi regions of Kurdistan. Kurdish in general and Sorani in particular, have been in contact with many regional languages, particularly Arabic, Persian, Assyrian, Turkish and Armenian. Therefore, the language is influenced in lexical borrowing, phonology and morphology [Chyet and Schwartz, 2003, Jügel, 2014]. Regarding Sorani dialectology, we proceed with the findings of the Dialects of Kurdish project [Matras, 2017a]. In this project which carries out a thorough analysis of Kurdish dialects based on audio samples of various Kurdish regions, Matras [Matras, 2017b] identifies two principal innovation zones within the Sorani-speaking regions of Iran and Iraq: Southern Sorani with Sulaymaniyah province as the epicenter and Northern Sorani with Erbil province as the epicenter. The innovations are specified based on structural differences, particularly in the morphology. Therefore, we believe that addressing Sorani Kurdish morphology from a formal point of view is beneficial to highlight the differences within Sorani subdialects more easily. For further details regarding Kurdish dialects, see the Database of Kurdish Dialects in the Database of Kurdish Dialects [Matras, 2017a].

### 2.2 Phonetics and Alphabet

Sorani Kurdish has eight vowels and 28 consonants [Thackston, 2006b]. Unlike the Kurmanji dialect for which the Latin script is widely used, Sorani dialect speakers favour the Persian-Arabic script. Using two scripts for writing a language scatters the readers and has been proven to be challenging in NLP tasks [Esmaili, 2012]. Although both orthographies in both scripts are phonemic, there is not always a direct mapping between the characters, particularly with respect to ‘و’ and ‘ی’ which can be transliterated into ‘w/u’ and ‘y/î’, respectively. On the other hand, the grapheme ”*i*”, which is known as *Bizroke* (literally meaning, “the little furtive”) among Kurdish linguists, does not have an equivalent in the Persian-Arabic orthography. [Ahmadi, 2019] presents a rule-based transliteration system that can successfully map ‘و’ and ‘ی’ but fails to detect Bizroke accurately. This is due to frequent usage of Bizroke within morphemes which is missing in writing. Therefore, with regards to morphology, absence of Birzoke in the Persian-Arabic orthography hinders automatic transliteration between morphemes containing this grapheme and further complicates the task of morphological analysis [Ahmadi and Hassani, 2020]. Table 1 provides the two most common scripts of Kurdish, Latin-based in the second row and Persian-Arabic in the third row along with the phonemes in International Phonetic Alphabet (IPA) in the first row.

| | | | | | | | | | | | | | | | | | | | | | | | | | | | | | | | | | | | | |
|---|---|---|---|---|---|---|---|---|---|---|---|---|---|---|---|---|---|---|---|---|---|---|---|---|---|---|---|---|---|---|---|---|---|---|---|---|
| aː | b | t͡ʃ | d͡ʒ | d | æ | eː | f | g | h | ɪ | iː | ʒ | k | l | ɫ | m | n | oː | p | q | ɾ | r | s | ʃ | t | ʊ | uː | v | w | x | j | z | ħ | ʕ | ɣ | ʔ |
| a | b | ç | c | d | e | ê | f | g | h | i | î | j | k | l | ł | m | n | o | p | q | r | ř | s | ş | t | u | û | v | w | x | y | z | ë | ḧ | ẍ | ’ |
| ا | ب | چ | ج | د | ە | ێ | ف | گ | ه | | ی | ژ | ک | ل | ڵ | م | ن | ۆ | پ | ق | ر | ڕ | س | ش | ت | و | وو | ڤ | و | خ | ی | ز | ع | ح | غ | ئ |

Table 1: A comparison of the two common scripts of Kurdish, Latin (second row) and Arabic-Persian (third row), along with the phonemes in IPA (first row)

### 2.3 Grammar

Sorani Kurdish is a null-subject language, that is a language where verbs can be used without pronouns. Kurdish has a subject-object-verb (S-O-V) order and can be distinguished from other Indo-Iranian languages by its ergative-absolutive alignment which appears in past tenses of transitive verbs. Some of the grammatical features of Kurdish, Zaza, Gorani, Persian and English languages in Table 2 based on [Ahmadi, 2020b].

A few features of the Sorani Kurdish syntax have been of interest to linguists, especially the syntax of prepositions [Samvelian, 2007a], person markers [Mohammadirad, 2020], Izafa constructions [Salehi, 2018] and verbal categories [Kareem, 2016].

### 2.4 Sorani Kurdish Data

Some of the interlinear glossed texts which are used in this paper are extracted from reference grammar books, namely [Thackston, 2006b, McCarus, 2007, Blau, 2000, Hacî Marif, 2000, McCarus, 1958]. In addition, the author relies on the linguistic knowledge as a native speaker of Sorani Kurdish.

## 3 Related Work

Kurdish morphology and syntax have been previously studied by many scholars. Excluding pedagogical resources describing general grammar of Sorani Kurdish, we can mention the following major contributions to the field. An overview of Kurdish linguistics is provided in [Haig and Matras, 2002].

| Language | Word order | Passive | Gender | Case | Alignment |
|---|---|---|---|---|---|
| Sorani Kurdish | S-O-V | morphological | (almost) no gender | nominative, oblique, locative, vocative | nominative–accusative, only in past transitive ergative–absolutive |
| Kurmanji Kurdish | S-O-V | periphrastic with *hatin* (to come) | feminine, masculine | nominative, oblique, Izafa, vocative | nominative–accusative, only in past transitive ergative–absolutive |
| Gorani | S-O-V | morphological | feminine, masculine | nominative, oblique, Izafa | nominative–accusative, only in past transitive ergative–absolutive |
| Zazaki | S-O-V | morphological | feminine, masculine | nominative, oblique, oblique of kinship terms, locative, vocative, double Izafe | nominative–accusative, only in past transitive ergative–absolutive |
| Persian | S-O-V | periphrastic with *shodan* (to become) | no gender | nominative, accusative, Izafa | nominative-accusative |
| English | S-V-O | periphrastic | no gender | nominative, oblique, genitive only for personal pronouns | nominative–accusative |

Table 2: A comparison of Sorani and Kurmanji dialects of Kurdish with Zazaki, Gorani, Persian and English languages

Among the resources written in Kurdish, [Hacî Marif, 2000] analyzes various morphological and syntactical forms in all Kurdish dialects. [Baban, 2001] focuses on the repetition suffix "*ewe*" in Sorani Kurdish and analyzes both morphology and syntax with respect to this morpheme.

There are also many contributions by non-native researchers. [Blau, 2000] and [McCarus, 1958] provide a descriptive analysis of Sorani Kurdish with a focus on three of the subdialects of Sorani, namely Mukriani, Babani and Ardalani. On the interaction of morphology and phonology, [Abdulla, 1980] and [Rashid Ahmad, 1987] analyze some aspect of the Kurdish phonology where various morphological rules are taken into account. One of the specific characteristics of Kurdish is ergatavity. This topic has been widely studied in Indo-Aryan languages [Karimi, 2012] in general, and Kurdish in particular [Bynon, 1980, Matras, 1997, Haig, 1998, Jukil, 2015, Mahalingappa, 2013, Karimi, 2010, Karimi, 2014]. Another interesting topic is the *Ezafe* construction which has been addressed as well [Karimi, 2007, Sahin, 2018].

Another related work to the current manuscript is [McCarus, 2007] which analyzes the inflectional and derivational patterns in the Sorani Kurdish morphology. [Fatah and Hamawand, 2014] provide descriptions of some of the Sorani Kurdish positive and negative prefixes based on the Prototype Theory. [Karacan and Khalid, 2016] analyze adjectives in various Kurdish dialects. [Haig, 2004] studies alignment in Kurdish from a diachronic perspective. [Walther, 2012] investigates morphological structure with respect to the Sorani Kurdish mobile person markers. This work extends the formal analysis of Sorani Kurdish endoclitic placement by [Samvelian, 2007b]. [Kolova, 2017] carries out a contrastive study on the Sorani Kurdish and English inflectional morphemes. [Doostan and Daneshpazhouh, 2019] studies *-râ* morpheme from syntactic and semantic perspectives.

Although the knowledge of morphology in creating NLP resources is of fundamental importance, there are only a limited number of studies that deal with Kurdish computational morphology, directly or explicitly within various tasks. [Salavati and Ahmadi, 2018] describe Kurdish morphology for the task of stemming and spelling error correction using $n$-gram models. They also provide a non-exhaustive list of the most frequent morphemes in Sorani Kurdish and report challenges in Sorani Kurdish lemmatization and spelling error correction due to morphological complexity. Walther and Sagot [Walther and Sagot, 2010] and Walther et al. [Walther et al., 2010] present a methodology and preliminary experiments on constructing a morphological lexicon for both Sorani and Kurmanji in which a lemma and a morphosyntactic tag are associated with each known form of the word. The lexicon[1] contains 22,990 word forms of 274 lemmata and is added to the Universal Morphology (UniMorph) project [2] which is a collaborative effort to improve how NLP handles complex morphology in the world's languages [Kirov et al., 2018, Cotterell et al., 2017]. [Gökırmak and Tyers, 2017] shed light on the morphology of Kurdish to create a treebank for Kurmanji. More recently, [Ahmadi, 2020a] creates a morphological analysis for the task of lexical analysis, also known as tokenization, for both Sorani and Kurmanji dialects of Kurdish.

# 4 Preliminary Definitions

Throughout this paper, we are using terms and notions which are introduced in this section.

A *word* is the most basic concept of morphology which can be defined as "*a contiguous sequence of letters*" [Haspelmath and Sims, 2013][3]. In morphology, words can be described by two fundamental notions: *word-forms* and *lexemes*. A word-form is the phonological sound or orthographic appearance of a word which can semantically identify something in a concrete way. A set of word-forms can belong to the same unit of meaning which is called a *lexeme*. In other words, different word-forms belonging to the same lexeme express the same core concept with different grammatical meanings. *Lemma* is another notion which refers to a more generic classification of word-forms and lexemes. That is, a canonical form is chosen by convention to represent a set of lexemes which is usually the least marked one and appears as the dictionary headword. In this work, we write lexemes in SMALL CAPITAL LETTERS and word-forms in *italics*.

**Example 4.1.** Three Sorani Kurdish lexeme families

(a) *dejîm* '(I) live', *jiyam* '(I) lived', *bijîmewe* 'that (I) live again', *jiyawim* '(I) have lived'

(b) *xwardim* '(I) ate', *xwardibûm* '(I) had eaten', *dexoyn* 'we eat', *bîxo* 'eat it!'

(c) *xwardiminewe* '(I) drank them', *detxwardewe* '(you) were drinking (it)', *bixorewe* '(you) drink!'

To further clarify the distinction between word-forms, lexemes and lemmata, Example 4.1 is provided with three lexeme families: (a) JIYAN 'to live', (b) XWARDIN 'to eat' and (c) XWARDINEWE 'to drink'. Various word-forms, such as *jiyam* and *bijîmewe* and many more, belong to the same lexeme JIYAN which expresses the same core concept of living. On the other hand,

[1] `https://github.com/unimorph/ckb`

[2] `https://unimorph.github.io`

[3] The definition of word is more challenging and can even be considered as a language-specific concept [Martin, 2017].

XWARDIN and XWARDINEWE are two different lexemes as they denote two distinct meanings. However, they have the same lemma which is *xwardin* and therefore, it is used as the headword in the dictionary for both of them. The task of finding the lemma of a word is called *lemmatisation* and is of importance in many NLP tasks. However, in morphology, our focus is on lexemes rather than lemmata since grammatical rules vary based on lexemes.

Word-forms are created following a set of morphological rules using morphemes. Morphemes are the smallest constituent of a linguistic expression that bear a concrete or abstract meaning [Haspelmath and Sims, 2013]. For instance, JIYAN represents the concrete meaning of the act of living, while *-ewe* in *bijîmewe* has an abstract meaning which can be described as the act of repeating something.

There are two types of morphological relationships: *inflectional* which studies the relationship between word-forms and lexemes, e.g. *dejîm* versus JIYAN, and, *derivational* which focuses on the relationship between lexemes of a word family, e.g. XWARDIN versus XWARDINEWE. In inflectional morphology, word-forms are composed of morphemes with concrete meanings and various morphemes with abstract meaning, which are called *affixes* and *clitics*. In Sorani Kurdish, affixes can follow or precede the main part of the word, in which case, they are called prefix and suffix, respectively. For instance, *dexoyn* is segmented into three morphemes, "*de + xo + yn*", where "*de-*" is a prefix and "*-yn*" is a suffix. We define the main part of the word to which affixes or clitics are attached as a stem or a base. Although stem and base are sometimes used interchangeably, we use stem for inflectional and base for derivational affixes. The base or stem that cannot be further analyzed into constituent morphemes is called a root.

Throughout this manuscript, we use the Leipzig glossing rules for interlinear morpheme-by-morpheme glosses [Comrie et al., 2008]. In order to facilitate the readability of the provided examples in Sorani Kurdish, we use its Latin-based script. In addition, the list of the Sorani Kurdish morphemes are provided in Appendix A in both Arabic-based and Latin-based scripts. It should also be noted that Sorani Kurdish has many sub-dialects in which the behaviour of some of the morphemes may vary. We mainly focus on morphological forms which are broadly accepted in such subdialects as the core morphology of the language. Variations, specifications and translations are also provided. If there are more than one form for a specific morpheme which is not an allomorph, we use "/" to specify them, e.g. "*î/ît*".

# 5 Sorani Kurdish Morphemes

Morphemes are classified into free and bound. While free morphemes are meaningful as they are, bound morphemes only carry meaning when affixed with other words. In this section, we provide a description of the Sorani Kurdish morphemes according to the classification provided in Figure 2. Although we cover all the bound morphemes in both derivational and inflectional forms, among the free morphemes adpositions are only presented.

Sorani Kurdish morphemes:
- Bound morphemes:
  - Clitics
    - Enclitics
    - Endoclitics
  - Affixes
    - Inflectional
    - Derivational
- Free morphemes:
  - Adpositions
  - Pronouns

Figure 2: A classification of Sorani Kurdish morphemes

## 5.1 Bound Morphemes

Affixes and clitics are similar in the way that they cannot constitute a word and they lean on a prosodic host, i.e. a word for stress assignment. On their differences, [Haspelmath and Sims, 2013] provides various tests to distinguish between affixes and clitics among which "*freedom of host selection*" is defined as the most salient property of clitics. Clitics can appear with hosts of various syntactic categories while affixes only combine with syntactically-related stems. Moreover, "*affix–base combinations may have an idiosyncratic meaning, whereas clitic–host combinations never do*" [Haspelmath and Sims, 2013, p. 199]. Following the convention, we mark the clitic and affix boundaries respectively with '=' and '-' signs.

The clitics and affixes in Sorani Kurdish have been widely studied previously and have been shown to be challenging considering the general theory of clitics [W. Smith, 2014, Gharib and Pye, 2018]. This problem is particularly observed with respect to the direct and oblique person markers which can appear in different positions within a word-form depending on the functionality. In this section, the clitics and affixes in Sorani Kurdish are described.

### 5.1.1 Clitics

Clitics are categorized based on their position with respect to the host. A clitic is called proclitic and enclitic, if it appears before and after the host, respectively. There are two other forms of clitics which are non-peripherical and exist only among a few number of natural languages. If a clitic appears between the host and another affix, it is called a mesoclitic. A different type of non-peripheral clitic is endoclitic which appears within the host itself and is unique to a few languages around the world, such as Udi [W. Smith, 2014], Pashto [Kopris, 2009], Degema [Kari, 2002], Andi [Maisak, 2018] and also Sorani Kurdish. Although [Anderson, 2005] considers both mesoclitic and endoclitic as synonym, they are also used separately to distinguish the intermorphemical and intramorphemical aspect of it. Following [Walther, 2012], we also use endoclitic to refer to both cliticization types as endoclitic.

**Personal markers and present copula** Sorani Kurdish personal markers have been previously a matter of extensive studies, particularly in [Bonami and Samvelian, 2008, Samvelian, 2007b, Bonami and Samvelian, 2008, Walther, 2012, W. Smith, 2014, Gharib and Pye, 2018, Jügel and Samvelian, 2020]. What makes Sorani personal endings intriguing for researchers is the pattern in which some of the clitics appear and morphotactic rules. Depending on their tense and transitivity, verbs use different endings for both subject, object, agent and patient marking. Such a distinction between personal markers in past and present tenses in transitive verbs is due to the split ergativity characteristic of Sorani [Haig, 2008, Akkus et al., 2019, Manzini et al., 2015]. We discuss the pattern of agreement in Sorani ergativity further below.

There are mainly four groups of personal markers in Sorani which are provided in Table 3 and described as follows:

- COP are enclitic personal endings based on the the copula BÛN (to be) serving as present copula.
- **PM**$_1$ are the personal markers for subject marking in present verbs of both transitive and intransitive verbs.
- **PM**$_2$ are the personal markers for subject and agent marking in imperative verbs.
- **PM**$_3$ are
  - the subject markers of intransitive verbs in past tenses
  - the patient markers of transitive verbs in past tenses
- **PM**$_4$ can be clitics and affixes. [Walther, 2012] defines these markers as mobile person markers (MPMs) according to the morphological reversal described in [Baerman, 2007]. It is an endoclitic due to its "... *seemingly erratic pattern*" in appearing within a verb form [Walther, 2012, p. 5]. The followings are the functions of these markers:
  - endoclitics as agent markers for transitive verbs in the past tenses
  - endoclitics as patient marker for transitive verbs in the present tenses
  - suffixes as possessive pronouns

| | Copula (COP) | Present | | Past | |
|---|---|---|---|---|---|
| | | ind & sbjv (PM$_1$) | imp (PM$_2$) | Intransitive (PM$_3$) | Transitive (PM$_4$) |
| 1SG | *=im* | *=im* | | *=im* | *=im* |
| 2SG | *=î/=ît* | *=î/=ît* | *=e* | *=î/=ît* | *=it* |
| 3SG | *=e* | *=ê/=êt* | | ∅ | *=î* |
| 1PL | *=în* | *=în* | | *=în* | *=man* |
| 2PL | *=in* | *=in* | *=in* | *=in* | *=tan* |
| 3PL | *=in* | *=in* | | *=in* | *=yan* |

Table 3: Personal markers in Sorani Kurdish

Although Sorani has a nominative-accusative alignment in present tenses, its ergative feature appears in the past tense, making it a split ergative language [Coon, 2013]. In past tenses, transitive verbs agree with the subject of intransitive (PM$_3$) and the agent of transitives (PM$_4$). In Example 5.1, a few examples are provided to further clarify the usage of the markers where PM$_1$, PM$_3$ and PM$_4$ are respectively represented in bold black, green and red. For instance, in Example 5.1.5, ”*man*” appears as an endoclitic for agent marking along with the zero suffix ∅ for patient marking while in Example 5.1.3, ”*yan*” of the same marker category of ”*man*”, appears as the object marker.

**Example 5.1.** Alignment in present and past tenses of KEWTIN (to fall) and GIRTIN (to get)

(1) *dekew**în***
*de-kew-**în***
fall.PRS.PROG.INTR.1PL
‘(**we**) are falling.’

(2) *degir**în***
*de-gir-**în***
get.PRS.PROG.TR.1PL
‘(we) are getting.’

(3) *de**yan**gir**în***
*de=**yan**-gir-**în***
get.PRS.PROG.TR.1PL.3PL
‘(**we**) are getting them.’

(4) *kewt**în***
*kewt-**în***
fall.PST.PROG.INTR.1PL
‘(**we**) fell.’

(5) *girt**man***
*girt=**man**-∅*
get.PST.PROG.TR.ERG.1PL.3SG
‘(we) got (it).’

(6) *girt**manin***
*girt=**man-in***
get.PST.PROG.TR.ERG.1PL.3SG
‘(we) got them.’

On the placement of clitics, [Walther, 2011, W. Smith, 2014] have previously analysed the case of Kurdish. To summarize, [W. Smith, 2014] notes that ”*clitics will attach to the leftmost element within the verbal phrase, and if there are no lexical elements to the left of the the verb within verbal phrase, then the clitic will attach to the leftmost morpheme within the verb itself*”. We complete this by adding that if there is no morpheme preceding the verb stem, then the agent marker, i.e. $PM_4$ in Table 3, appears as an endoclitic in different placements according to other clitics and suffixes. As a general rule in most subdialects of Sorani, the agent marker usually appears before the patient marker as in Example 5.1.6.

Table 4 provides an example with the verb GIRTIN (to take, to get) in the simple past tense. In Table 5 the same verb is provided with all clitics and suffixes that can possibly emerge in a verb form. We see that the endoclitic *im* as the agent marker appears before the stem once a morpheme appears before the stem and remains in the second position within the verb form as long as a preceding morpheme exists and no other clitic appears in the verb form. In Table 5 we see that the placement of the agent marker changes when the endoclitic *=îş* appears. We discuss further below how the endoclitic *=îş* changes the placement of this marker.

| | | | | | | | | | | |
|---|---|---|---|---|---|---|---|---|---|---|
| 0 | | | | | girt | | | | | past stem of GIRTIN (to take, to get) |
| 1 | | | | | girt | im | | | | I got |
| 2 | | | | | girt | im | in | | | I got them |
| 3 | | | | | girt | im | in | e | | I got them to/with |
| 4 | | | | | girt | im | in | e | ewe | I got them to/with again |
| 5 | | | ne | im | girt | in | e | ewe | | I did not get them to/with again |
| 6 | | ne | im | de | girt | in | e | ewe | | I was not getting them to/with again |
| 7 | heł | im | ne | de | girt | in | e | ewe | | I was not lifting them to/with again |

Table 4: The placement of the agent marker (in grey boxes) with respect to the base in a verb form

| | | | | | | | | | | | | |
|---|---|---|---|---|---|---|---|---|---|---|---|---|
| 0 | | | | | | girt | | | | | | past stem of GIRTIN (to take, to get) |
| 1 | | | | | | girt | im | | | | | I got |
| 2 | | | | | | girt | im | in | | | | I got them |
| 3 | | | | | | girt | im | in | e | | | I got them to/with |
| 4 | | | | | | girt | im | in | e | ewe | | I got them to/with again |
| 5 | | | | | | girt | îş | im | in | e | ewe | I got them also to/with again |
| 6 | | | ne | îş | im | girt | in | e | ewe | | | I did not get them also to/with again |
| 7 | | ne | îş | im | de | girt | in | e | ewe | | | I was not getting them also to/with again |
| 8 | heł | îş | im | ne | de | girt | in | e | ewe | | | I was not lifting them also to/with again |

Table 5: The placement of the endoclitic *=îş* (in green boxes) and agent marker (in grey boxes) with respect to the base and each other in a verb form

**Enclitic/Endoclitic *=îş*** is an enclitic and endoclitic, based on the word structure, meaning *also*, *even* and *too* which can be added to all parts of speech, particularly nouns, noun–adjective phrases, pronouns and verbs in specific moods[4]. In all part of speeches except verbs, *=îş* appears as an enclitic at the end of the word form. However, when it appears in a verb form, it follows an erratic pattern similar to $PE_4$. In cases where the two endoclitics appear together, like in *hełîşimnedegirtinewe* ((I) didn't lift them again) in Table 4, *=îş* precedes the agent marker $PE_4$. To summarize, the presence of a leftmost morpheme and other clitics determine the placement of the endoclitics.

Example 5.2 provides a few word forms where *=îş* appears. If a lexical element exists to the left of the verb, like "*kitêbêk*" (a book) or "*min*" (I, me), *=îş* sticks to them as en endoclitic. Otherwise, *=îş* attaches to the leftmost morpheme, as in *heł-* and *he-* affixes in *hełîşhatim* and *heşimban*. Consider the verb form *hatîşim* ((I) came) where the derivational morpheme *heł* is removed, the enclitic appears before the personal ending. Similarly, in Example 5.2.4, *=îş* appears after the first morpheme *he* of the verb HEBÛN (to have).

**Example 5.2.** Enclitic/Endoclitic *=îş*

(1) *he**şim**ban.*
*he=îş-im-ba-in.*
have.PST.1SG.3PL also
'(even if I) have had them'

(2) *hat**îş**im.*
*hat=îş-im.*
come=also-PST.1SG.
'(I) also came'

(3) *heł**îş**hatim.*
*hel=îş-hat-im.*
up=also-flee-PST.1SG.
'(I) also fled away'

(4) *kitêbêk**îş** heye.*
*kitêb-êk=îs heye.*
book-INDEF=also COP.PRES.3SG.
'There is also a book.'

(5) *min**îş** betenê hełhatim.*
*min=îş be-tenê hel-hat-im.*
I-NOM=also alone.ADV up.flee-PST.1SG.
'I also fled away alone' (emphasis on the act)

(6) *min beten**êş** hełhatim.*
*min=îş be-tenê=ş hel-hat-im.*
I-NOM alone.ADV.even up.flee-PST.1SG.
'I fled away alone too' (emphasis on alone)

Having said that, the position of the enclitic *=îş* may slightly vary among Sorani Kurdish subdialects. In the Ardalani subdialect, for instance, *=îş* mostly appears as an enclitic as in *hełhatim**îş*** versus *heł**îş**hatim* explained above.

[4] [Thackston, 2006b] believes that "*it cannot follow a finite verb form*" which does not seem to be the case.

**Example 5.3.** Directional complement *=e* and pronominal adverb *=ê*

(1) *geyîştime małyan.*
*geyîşt-im=e mał-yan.*
arrive-PST.PPFV.1SG-to house-PM4.
'(I) arrived/have arrived to their house.'

(2) *geyîştimê.*
*geyîşt-im=ê.*
arrive-PST.PPFV.1SG-thereto.
'(I) arrived/have arrive there.'

(3) *kitêbeke be to dedem.*
*kitêb-eke be to de-de-m.*
book.SG.DEF to you.2SG.OBL give.PRS.PROG.1SG.
'(I) give the book to you'

(4) *kitêbeket pê dedem.*
*kitêb-eke-t pê de-de-m.*
book.SG.DEF to give.PRS.PROG.1SG.
'(I) give you the book'

(5) *kitêbeket dedemê.*
*kitêb-eke=t de-de-m=ê.*
book.SG.DEF-ART.SG=PE4 give.PRS.PROG.1SG.
'(I) give you the book'

(6) *detdemê.*
*de=t-de-m=ê.*
give.PRS.PROG.1SG.you.sg.2nd.acc
'(I) give you that'

**Postposed directional complement =*e*** is used to indicate a directional aspect in the verb act, whether as a movement or a figurative form to represent meaning of to or for.

**Pronominal adverb =*ê*** In sentences with phrasal verbs with a preposition or a postposed directional complement, the pronominal adverb *ê* replaces the antecedent prepositional object, i.e. *be* and -*e*, oblique pronoun, e.g. *wî* 'him/her (oblique)', accusative nouns or locative adverb, e.g. *ewê* 'there'.

Example 5.3 provide a few sentences where the directional complement -*e* and pronominal adverb =*ê* are respectively specified in red and blue. In 5.3.2, =*ê* replaces the locative adverb *małyan* 'their house' with =*ê*. On the other hand, in 5.3.5 the enclitic =*ê* replaces the absolute preposition *pê* and in 5.3.6, it replaces the accusative noun *kitêbeke* 'the book' and its indirect object -*t* 'you'. 5.3.3 and 5.3.4 are useful to understand how the simple preposition *be* appears in its absolute form *pê* with a change in the position of the indirect object -*t* 'you'.

### 5.1.2 Affixes

| Nouns | Verbs | Adjectives | Adverbs |
|---|---|---|---|
| **number** (SG, PL) | **number** (SG, PL) | **number** (SG, PL) | **degree** (COMP, SUPL) |
| **person** (1, 2, 3) | **person** (1, 2, 3) | **degree** (COMP, SUPL) | |
| **determiners** (DEF, IND, DEM) | **mood** (IND, SBJV, IMP, COND) | **determiners** (DEF, IND, DEM) | |
| **case** (OBL, LOC, VOC) | **aspect** (PRF, IMP, PROG) | | |
| **gender** (M, F) | **tense** (PST, PRS) | | |

Table 6: Inflectional features and values of Sorani Kurdish. It should be noted that the function of cases and genders vary among Sorani subdialects and are not widely in use.

In addition to clitics, there are many affixes in Sorani Kurdish which are presented in this section. Our focus is on the most frequent affixes belonging to open-class parts of speech, namely nouns, verbs, adjectives and adverbs, and which are used in inflectional (as shown in Table 6) and derivational paradigms.

### 5.1.3 Inflectional Affixes

**Affixes for nouns** Table 7 presents the suffixes that appears with nouns based on their determiners, i.e. absolute, indefinite, definite and demonstrative, and also their cases, i.e. oblique, vocative and locative. It should be noted that the usage of cases and genders among subdialects of Sorani Kurdish varies based on the regions. Generally speaking, cases and genders are more respected in the Northern Sorani innovation zone, while the Southern Sorani zone ignores the cases oftentimes. For more information, see the distribution of genders and case marking see the Kurdish Dialect Project[Matras, 2017c][5].

The usage of the plural suffixes vary depending on the subdialects and also borrowings. In the Kurdish regions of Iran, under the influence of Persian, the enclitic *-ha* (ها) is sometimes in plural forms of words ending with a vowel, e.g. *cêgeha* (places.IND.PL) and also, with some Persian loanwords, e.g. *danişgaha* (universities.IND.PL from the Persian borrowing "*daneshgah*" (دانشگاه, 'university'), but also with some Kurdish words. However, this practice is usually avoided in writing. Moreover, the plural suffix *-at* from Arabic is used for plural form of etymologically Arabic words such as *małat* (ماڵات, 'cattle', domestic animals) which comes from the Arabic word *mał* ماڵ, 'possession', 'wealth').

| | | | |
|---|---|---|---|
| determiner | absolute | singular | ∅ |
| | | plural | ∅ |
| | indefinite | singular | *-êk* |
| | | plural | *-an, -gel, -ha, -at* |
| | definite | singular | *-eke* |
| | | plural | *-ekan* |
| | demonstrative | singular | *-e* |
| | | plural | *-ane, -gele* |
| case | nominative | singular/plural | ∅ |
| | oblique | masculine | *-î* |
| | | feminine | *-ê* |
| | vocative | masculine | *-e* |
| | | feminine | *-ê* |
| | | plural | *-îne* |
| | locative | absolute | *-ê* |

Table 7: Suffixes associated with nouns in Sorani Kurdish

**The Izafa particle** (EZF) *Izafa*, a word originally from Arabic literally meaning "addition", refers to the unstressed grammatical particle *î* which appears between a head and its dependents in a noun phrase. Izafa is mainly used in possessive constructions where it appears between two parts, such as "*kitêbî min*" (book of me, my book) and can be translated as 'of' [Thackston, 2006b]. Moreover, in attributive adjectives where a noun as the head is followed by an adjective, i.e. noun + î + adjective, the Izafa construction varies based on the type of the head as follows:

- If the noun is absolute singular, indefinite singular or indefinite plural, the Izafa particle appears after the noun as *î*. This is described as open Izafa noun-adjective construction by [Thackston, 2006b].
- If the noun is a demonstrative or a definite, the Izafa particle changes from *î* to *e*. This is called close Izafa noun-adjective construction by [Thackston, 2006b].

[5] `http://kurdish.humanities.manchester.ac.uk/determiner-masculine-direct-object/`

[Salehi, 2018, p 53] demonstrates that the "*Izafa exhibits properties of a morphological expression that is contextually induced and is sensitive to properties that relate to definiteness*" and therefore, states that the =*e* enclitic in demonstrative and definite forms is in fact an allomorph of the marker =*eke*. Example 5.4 provides examples for each of these constructions.

| | |
|---|---|
| construct | -*î* |
| linker | *î, hî, hîn* |
| compounding | -*e* |

Table 8: The Izafa forms in Sorani Kurdish [Gutman, 2016, p 253]

**Example 5.4.** Izafa construction in Sorani

(1) *gulî min*
*gul î min*
flower.ABS-EZF me.OBL
'my flower'

(2) *gulî ciwan*
*gul î ciwan*
flower.ABS-EZF beautiful.ADJ
'beautiful flower'

(3) *gulêkî ciwan*
*gul-êk î ciwan*
flower=IND.SG-EZF beautiful.ADJ
'a beautiful flower'

(4) *gule ciwaneke*
*gul-e ciwan-eke*
flower.DEF-EZF beautiful.ADJ=DEF.SG
'the beautiful flower'

(5) ew *gule ciwane*
ew *gul-e ciwan-e*
that flower.N.DEF-EZF beautiful.ADJ=DEM.SG
'that beautiful flower'

(6) ew *gule ciwanane*
ew *gul-e ciwan-ane*
that flower.N.DEF-EZF beautiful.ADJ=DEM.PL
'those beautiful flowers'

There is another Izafa noun-adjective construction in Sorani Kurdish which is used to create nominal locutions by reversing the order of the noun and the adjective in the form of "adjective + e + noun". In this case, the noun appears in the absolute form and the modified form of the Izafa particle *e* is used. Although, such a structure is literally equivalent to the original construction of "noun + î + adjective", a different sense could be constructed this way. For instance, in the following example, "*xase kew*" and "*kewî xas*" are referring to the same things in literal meaning, however, that means a 'chukar partridge' while the latter is 'partridge'.

**Example 5.5.** Adjective-noun Izafa construction in Sorani

(1) *xase kew*
*xas-e kew*
good.ADJ-EZF partridge.ABS
'chukar partridge'

(2) *kele pyaw*
*kel-e pyaw*
big.EZF man.ABS
'nobleman'

**Affixes for adjectives and adverbs** There are two types of attributive adjectives in Sorani Kurdish:

- Postpositive adjectives are the most frequent types in Sorani Kurdish which appear after the noun in one of the following constructions:
  - with no additional morpheme, i.e. [noun + adjective], as in *ser zil* "head big" (big head)

- or within an Izafa construction, i.e. [noun + î + adjective] as in *guł î ciwan* (beautiful flower). [Thackston, 2006b] refers to this as open adjectival Izafa.

- Prepositive adjectives are those adjectives which appear before the noun in the Izafa construction of [adjective + *e* + noun] with *e* as the allomorph of the Izafa particle *î*. For instance, *sûr* 'red' and *guł* construct an adjective phrase as *sûre guł* (red flower). Prepositive adjectives are less frequently used and may modify the adjective phrase semantically too.

Similar to nouns, adjectives can be inflected based on definiteness and plurality using the same morphemes introduced in Tabel 7. In addition, the comparative and superlative forms of adjectives, and also adverbs, is expressed morphologically. For comparative and superlative forms, suffix *-tir* and *-tirîn* are respectively used.

**Example 5.6.** Adjective and adverb comparison forms in Sorani

| BASE | COMP | SUP | GLOSS |
|---|---|---|---|
| *nwê* | *nwêtir* | *nwêtirîn* | new |
| *ciwan* | *ciwantir* | *ciwantirîn* | beautiful, young |
| *ciwanane* | *ciwananetir* | *ciwananetirîn* | beautifully, youthfully |

**Modal prefixes** are used as verbal prefixes to demonstrate mood and aspect as follows:

- ***de-***: progressive marker in past and present tenses
- ***e-***: another progressive marker in other subdialects (particularly, Sulaymaniyah and Sanandaj subdialects)
- ***bi-/we-***: subjunctive and imperative
- ***ne-***: negation in past and present tenses, negative subjunctive
- ***na-***: negation only in present tenses
- ***me-***: negative imperative

| Voice | Transitivity | Verb | | | |
|---|---|---|---|---|---|
| | | Infinitive | Past stem | Present stem | Root |
| Active | Transitive | KÊŁAN (to plough) | *kêła* | *kêł* | *kêł* |
| | | BIRDIN (to bring) | *bird* | *be* | *bi* |
| | | HÊNAN (to carry) | *hêna* | *hên* | *hên* |
| | | BIRJANDIN (to roast) | *birjand* | *birjên* | *birj* |
| | | SÛTANDIN (to burn) | *sûtand* | *sûtên* | *sût* |
| | Intransitive | ÇÛN (to go) | *çû* | *çi* | *çi* |
| | | GEYŞTIN (to arrive) | *geyşt* | *ge* | *gi* |
| | | BÛN (to be) | *bû* | *bi* | *bi* |
| | | MAN (to stay, to remain) | *ma* | *mên* | *mi* |
| | | MIRDIN (to die) | *mird* | *mir* | *mir* |
| Passive | Intransitive | KÊŁDIRAN (to be ploughed) | *kêłdira* | *kêłdirê* | *kêł* |
| | | BIRDIRAN (to be brought) | *birdira* | *birdirê* | *bi* |
| | | HÊNDIRAN (to be carried) | *hêndira* | *hêndirê* | *hên* |
| | | BIRJÊNDIRAN (to be roasted) | *birjêndira* | *birjêndirê* | *birj* |
| | | SÛTÊNDIRAN (to be burnt) | *sûtêndira* | *sûtêndirê* | *sût* |

Table 9: Infinitive, past stem and present stem of a few Sorani Kurdish verbs

### 5.1.4 Verbal Derivational Affixes

**Infinitive suffix *-in*** The infinitive form of the verbs in Sorani Kurdish, similar to Kurmanji, ends with *-in* suffix. The base of the infinitive is the past stem which along with the present stem are used for constructing verb forms. The present stem, unlike the past stem, is a weak suppletive allomorph as it exhibits some similarity which cannot be described by phonological rules. A few infinitive verbs with past and present stems are provided in Table 9.

**Rule 5.1.** Rules of deriving past and present stems from infinitive

$$\begin{bmatrix} /XY(\text{in})/_V \\ x(= \text{INFINITIVE}) \end{bmatrix} \leftrightarrow \begin{bmatrix} /XY/_V \\ \text{PAST STEM OF } x \end{bmatrix} \leftrightarrow \begin{bmatrix} /XZ/_V \\ \text{PRESENT STEM OF } x \end{bmatrix}$$

The link between the past and present stems with the infinitive form of a verb is defined in Rule 5.1 where $X$ is the verb root, $Y$ and $Z$ are derivational suffixes as follows:

- **Past stem derivational suffixes**: *∅, -a, -and, -ard, -d, -î, -îşt, -ird, -t, -û, -y*
- **Present stem derivational suffixes**: *∅, -e, -ê, -ên, -êr*

For instance, by removing the *-in* suffix in the infinitive form of the verb SÛTANDIN in Table 9, the past stem *sûtand* is extracted. The present stem of the same verb *sûtên*, however, is a weak suppletive allomorph of the past stem without any explainable morphological change. On the other hand, if we start with the root of the verb, i.e. *sût*, and attach *-and* and *-ên*, the past and the present stems are created respectively.

**Passive voice derivational suffix *-diran*** Kurmanji dialect of Kurdish constructs passive voice by creating a compound where the transitive verb is accompanied by the verb "*hatin*" (to come) [Thackston, 2006a]. For instance, *hate xwarin* (was eaten) is the passive voice of *xwarin* (to eat) where the infinitive has not changed but *hatin* is conjugated. However, in the Sorani dialect, there is a relatively more complex process for passive marking.

The passive form of verbs in Sorani is created by appending *diran* or its allomorph *ran*, depending on the subdialect, to the present stem of the transitive verb. Table 9 provides the passive form of the transitive verbs. For instance, the infinitive HÊNDIRAN is created by adding *diran* to *hên*, the present stem of HÊNAN. Although this rule is mostly correct to generate passive form of transitive verbs in Sorani, there are a few examples where the passive is not morphologically derived from the same root but is another lexeme. [Thackston, 2006b] provides a list of such common verbs with irregular passive forms.

**Verbal derivational suffixes** In comparison to the number of the verbs in Sorani, a large proportion are derived from single-word verb lexemes [Walther and Sagot, 2010]. We can categorize the following processes through which Sorani verbs are extended based on the initial lexemes:

1. **Compound verbs**
   (a) **Open-class compounds** which use open-class words along with a single-word verb, particularly of (Noun + Verb) and (Adjective + Verb) structure. KIRDIN (to do) and BÛN (to be) are frequently used for this purpose.
   (b) **Particle verbs** use particles as demonstrated in Table 11 along with single-word verbs

(c) **Adpositional verbs** are another type of compound verbs similar to particle verbs where an adposition is used along with a single-word verb. A list of the Sorani Kurdish adpositions is provided in Table 14.

2. **Morphologically-derived verbs** refer to those single-word verbs which are created by the use of derivational affixes. For instance, the verb KEWTIN (to fall.INF) with the root *kew* can have another form by using the *-andin* derivational suffix to create KEWANDIN (to cause to fall, to drop). Table 10 provides some of the major derivational suffix for verb formation in Sorani Kurdish.

| | | |
|---|---|---|
| **I. Deverbal verbs** (V → V) | | |
| causative verb | ***mir****din* 'to die' | ***mir****andin* 'to kill' |
| applicative verb | ***kir****din* 'to do' | ***kir****diran* 'to be done' |
| anticausative verb | ***biř****în* 'to cut' (tr.) | ***biř****an* 'to cut' (intr.) |
| repetitive verb | ***xwar****din* 'to eat' | ***xwar****dinewe* 'to drink' |
| reciprocal verb | ***ał****an* 'to twist' | *lêk* ***ał****an* 'to twist together' (intr.) |
| **II. Denominal verbs** (N → V) | | |
| 'do the action of N' | ***fikir*** 'thought' | *tê* ***fikir****în* 'to think' |
| | ***nirx*** 'price' | ***nirx****andin* 'to price, to evaluate' |
| **III. Denominal nouns** (N → N) | | |
| diminutive noun | ***kiç*** 'girl' | ***kiç****ole* 'little girl' |
| inhabitant noun | ***jiwan*** 'tryst' | ***jiwan****ge* 'tryst place' |
| agent noun | ***asin*** 'iron' | ***asin****ger* 'blacksmith' |
| status noun | ***bira*** 'brother' | ***bira****(y)etî* 'brotherhood' |
| **IV. Denominal adjectives** (N → A) | | |
| diminutive noun | ***quř*** 'mud' | ***quř****awî* 'muddy' |
| proprietive (= 'having N') | ***aw*** 'water' | ***aw****î* 'watery' |
| material | ***çerm*** 'leather' | ***çerm****în* 'leathery' |
| characteristic | ***tirs*** 'fear' | ***tirs****in* 'coward' |
| augmentative | ***şiř*** 'ragged' | ***şiř****oł* 'very ragged' |
| **IV. Deadjectival noun** (A → N) | | |
| facilitative | ***sard*** 'cold' | ***sard****emenî* 'beverage' |
| **IV. Deadjectival adjectives** (A → A) | | |
| negative | ***xoş*** 'agreeable' | *na****xoş*** 'unagreeable' |
| endearing | ***xirt*** 'round' | ***xirt****ołe* 'chubby' |
| attenuative | ***řeş*** 'black' | ***řeş****baw* 'blackish' |

Table 10: Common derivational meanings of verbs in Sorani Kurdish. Roots are bold in black and derivational suffixes are specified in green

For instance, GIRTIN (to get, to take) as a single word can create new senses such as in "*dagirtin*" (to take down), "*bergirtin*" (to take ahead; to prevent) and "*masî girtin*" (to take fish; to fish). Although the Noun + Verb compound verbs are written separately or connected, depending on the orthography, particle verbs are always written together. This makes the particle behave like verbal prefixes. Figure 3 shows the morphological trees of few derived verbs from KIRDIN (to do).

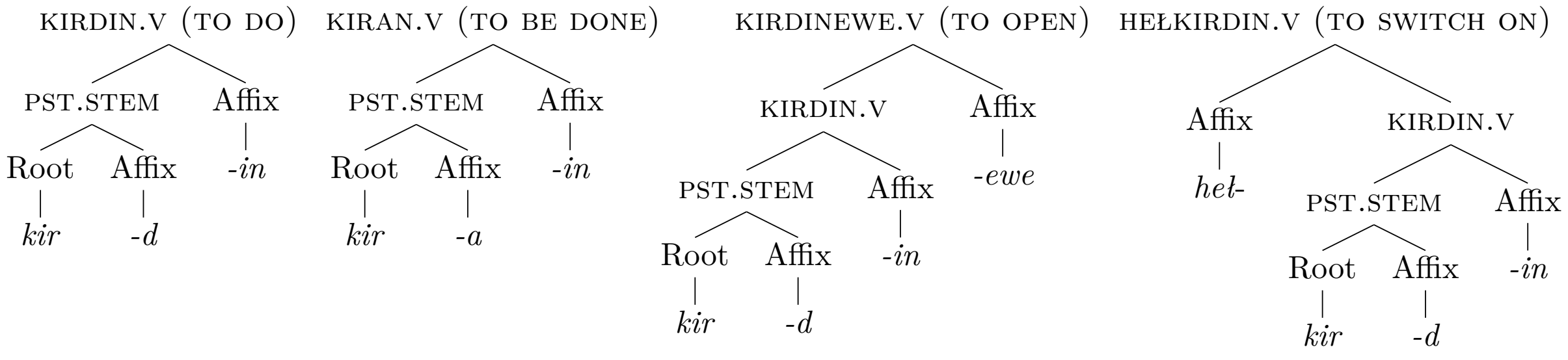

Figure 3: Morphological trees of KIRDIN (to do) and a few other derived verb forms

**Verbal suffix *-ewe*** In addition to a postposition (Table 14), *-ewe* can have the following roles:

- Inflectional verbal suffix for attributing the act of the repetition of the verb, e.g. *sûtandinewe* (to burn again)
- Derivational verbal suffix as in *kirdinewe* (to open) vs. KIRDIN (to do) and *xwardinewe* (to drink) vs XWARDIN (to eat)

In any case, if a verb appears with the postposed directional complement *-e* and *-ewe*, following a morphophonological rule, the combination of *-e* and *-ewe* remains *-ewe*.

**Verbal particles** Verbal particles are different from prepositions as they cannot take any complement. They are widely used for deriving new lexemes from single-word word forms in Sorani Kurdish. Table 11 presents the verbal particles in Sorani and their frequent semantic connotations when used with a verb.

| derivational prefix | description | example |
|---|---|---|
| *ber-* | ahead | *bergirtin* ‘to prevent’ |
| *da-* | down | *dagirtin* ‘to take down’ |
| *der-* | out | *derhênan* ‘to bring out’ |
| *heł-* | up | *helgirtin* ‘to pick up’ |
| *ser-* | over | *sergirtin* ‘to cover’ |
| *řa-* | towards, with | *řagirtin* ‘to retain’ |
| *řo-* | above | *řonîştin* ‘to sit down’ |
| *wer-* | upside, trans- | *wergirtin* ‘to receive’ |

Table 11: Derivational Verbal Particles with lexemes GIRTIN (to get, to take) and

## 5.2 Free Morphemes

Among the free morphemes, we present adpositions and pronouns as follows:

**Pronouns** Tables 12 and 13 provides pronouns in the dialects of Northern and Southern Sorani[6]. Although what is recognized as standard Sorani does not contain any grammatical case, some of the subdialects of Sorani Kurdish still use cases, particularly the Mukriani dialect spoken in the Kurdish regions of West Azerbaijan province in northwestern Iran and Hewlêrî dialect spoken in Erbil and its surrounding region.

[6]Based on [Matras, 2017b]’s classification of Sorani.

| Person | Independent | | Affixed | |
|---|---|---|---|---|
| 1SG | *min* | من | *-im* | م- |
| 2SG | *to* | تۆ | *-it* | ت- |
| 3SG | *ew* | ئەو | *-î* | ی- |
| 1PL | *ême* | ئێمە | *-man* | مان- |
| 2PL | *êwe* | ئێوە | *-tan* | تان- |
| 3PL | *ewan* | ئەوان | *-yan* | یان- |

Table 12: Independent and affixed pronouns in Southern Sorani

| Person | Nominative | | Other cases | |
|---|---|---|---|---|
| 1SG | *emin* | ئەمن | *min* | من |
| 2SG | *eto* | ئەتۆ | *to* | تۆ |
| 3SG | *ew* | ئەو | *wî* | وی |
| 1PL | *ême/eme* | ئێمە/ئەمە | *me* | مە |
| 2PL | *engo* | ئەنگۆ | *ingo* | نگۆ |
| 3PL | *ewan* | ئەوان | *wan* | وان |

Table 13: Independent pronouns in Northern Sorani dialects. Unlike this, the distinction between nominative and non-nominative cases does not exist in Southern Kurdish dialects.

**Adpositions** Adpositions are free morphemes which can be categoried as prepositions, postpositions and circumpositions.

Sorani Kurdish has a rich class of prepositions which are classified into primary prepositions and non-primary prepositions [Samvelian, 2007a]. Some of the primary prepositions, in addition to the simple form, have an absolute form which bears lexical stress and appears with different syntactic behaviour [McCarus, 1958, Samvelian, 2007a]. Table 14 lists the primary prepositions of Sorani Kurdish. On the other hand, non-primary prepositions are compound prepositions where a primary preposition is used with other grammatical elements, such as substantives, e.g. *le ser* ‘over’. Furthermore, postpositions can appear with prepositions to form circumpositions, e.g. *le ... da* (in).

It should be noted that the classification of adpositions in Kurdish ave been studied from various other perspectives, such as [Saeed, 2017].

| | simple | absolute | gloss |
|---|---|---|---|
| prepositions | *be* | *pê* | with, by, to |
| | *bê* | | without |
| | *bo* | *bo* | to, for, towards |
| | *de* | *tê* | at, in |
| | *le* | *lê* | from, in |
| | *legeł, degeł, letek* | *legeł, degeł, letek* | with |
| | *ře* | | with |
| | *we* | *wê* | to |
| postpositions | *da/a* | | in |
| | *řa* | | from |
| | *ewe* | | from |

Table 14: Adpositions in Sorani Kurdish

In addition to the absolute forms, prepositions can appear in a reciprocal form using cardinal number *yek* ‘one’ in Sorani, as presented in Table 15. . When used with an absolute preposition, the cardinal number appears with its allomorph *k*, as in *pêk* ‘to each other’. The absolute reciprocal prepositions are widely used to derive verbs.

| simple reciprocal | absolute reciprocal | gloss |
|---|---|---|
| *le yek* | *lêk* | from each other |
| *be yek* | *pêk* | to each other |
| *de yek* | *têk* | in each other |
| *we yek* | *wêk* | to/with each other |

Table 15: Absolute reciprocal prepositions in Sorani Kurdish

# 6 Morphophonology of Sorani Kurdish

Morphemes often have different phonological forms depending on the structure of the word and the surrounding morphemes. Sorani Kurdish clitics and affixes often exhibit allomorphy, that is having more than one single shape for the same morpheme. For instance, the pronominal endoclitic *=im* in *xwardim* '(I) ate' has another form following the phonological conditioning, which is *m*, as in *dejîm* '(I) live'. Such allomorphs can be reproduced following two types of alternations: automatic and morphophonological.

Automatic alternations are those alternations which can be described in phonological terms. On the other hand, morphophonological alternations occur under specific morphological constraints. These two alternations constitute the underlying representation of the words. What is pronounced by a native speaker, also known as the surface representation, is the product of applying alternation rules which are described in this section for Sorani Kurdish.

## 6.1 Automatic Alternations

The automatic alternation rules in Sorani Kurdish follow globally two rules which are consisted of using semi-vowels ”y” [j] and ”w” [w] wherever stringing two morphemes brings two vowels together and also eliding vowels. The choice of one of those semi-vowels may vary across Sorani sub-dialects. Some of the most frequent alternations are the followings:

**Rule 6.1.** Automatic alternation rules for Sorani Kurdish

(a) When two identical vowels come together, one of them elides as in $e + e \rightarrow e$ and $a + a \rightarrow a$
Example: *şêwe* 'manner' + *e* + *ciyawaz=eke* (different.ADJ.SG.DEF) → *şêw**ee** ciyawazeke* → *şêwe ciyawazeke* "the different manner"

(b) The short vowel *i* elides whenever appears with another vowel, as in $e + i \rightarrow e$ and $a + i \rightarrow a$
Example: *sawa* (toddler) + *-eke* → *saw**ae**ke* → *sawake* "the toddler"

(c) When two different vowels come together, depending on the vowels, semi-vowels *y* or *w* may be used:
Examples: *name* (letter) + *-êk* → *nameyêk* "a letter" / *name* (letter) + *=îş* → *nameyş* "also letter"

(d) /XV:CiC/ + *-V(C)* (a morpheme starting with a vowel) → /XV:CCVC/
Example: *birdûmin* "(I) have taken them" + *-e* → *birdûmne* "(I) have taken them to"

## 6.2 Morphophonological Alternations

In addition to the phonological allomorphs which have rather similar pronunciation shapes and bear the same meaning, there are various cases where morphology adds auxiliary morphemes during the morphological process. Such alternations are morphophonological and can be described as follows where X, C and V respectively refer to a word, a consonant and a vowel:

**Rule 6.2.** Morphophonological alternation rules for Sorani Kurdish

(a) X (V.PRS.PRF) + =*e* (directional suffix) → X*te*
***=te***: allomorph of =*e* (directional suffix)
Example: *birdûme* "(I) have taken" + =*e* → *birdûmete* "(I) have taken to"

(b) X (V.IMP.2SG) + -*ewe* (repetition suffix) → X*rewe*
***-rewe***: allomorph of -*ewe* (repetition suffix)
Example: *bixwêne* (read.V.IMP.2SG) + -*ewe* → *bixwênerewe* (read.V.IMP.2SG again)

(c) /XCæ/ (V.PRS.STEM) + =*ê*/=*êt* (3SG present personal marker) → /XCaː/
***=a/=at***: allomorph of =*ê*/=*êt* (3sg pronominal enclitic)
Example: *debe* + =*ê* → *deba* "(he/she/it) takes"

(d) /Xoː/ (V.PRS.STEM) + =*ê*/=*êt* (3SG present personal marker) → /X*w*aː/
***=wa/=wat***: allomorph of =*ê*/=*êt* (3sg pronominal enclitic)
Example: *dexo* + =*ê* → *dexwa* "(he/she/it) eats"

(e) XV (noun) + =*îş* (emphasis endoclitic) → XV*ş*
***=ş***: allomorph of =*îş*
Example: *name* (letter) + =*îş* → *nameş* "also letter"

(f) XV (noun) + -*êk* (*ish*) → XV*yek*
***-yek***: allomorph of -*êk* (singular indefinite suffix)
Example: *name* (letter) + -*êk* → *nameyek* "a letter"

# 7 Morphological Rules

Morphological patterns can be defined as rules. There are mainly two approaches in writing morphological rules: the morpheme-based model and the word-base model [Haspelmath and Sims, 2013, p. 40]. Morpheme-based model describes how the combination of morphemes in a language can create new word forms, while the word-base model "formulates word-schemas that represent the features common to morphologically-related words" [Haspelmath and Sims, 2013, p. 46] rather than splitting them into parts.

In this section[7], we use the morpheme-based model to define word-structure rules for Sorani Kurdish. We believe that this model is ideal to represent the procedural nature of morphological constructions which can be used to create morphological analyzers using computational methods. We describe the rules for the open-class parts of speech, mainly adjective, adverb, preposition, noun and verb. Following these rules, we present some of the major morphophonological features which create the surface realisation of word forms based on the phonology of Sorani Kurdish.

Throughout the morphological rules, + refers to the concatenation of the strings sequentially. If the presence of a specific morpheme within a construction is possible, parentheses are used. For instance, the following description

- (derivational prefix +) (+ inflectional prefix +) + past stem + inflectional suffix (+ derivational suffix)

[7] Rules regarding other open-class open class parts of speech will be added as well.

can produce the following rules:

- past stem + inflectional suffix
- past stem + inflectional suffix + derivational suffix
- inflectional prefix + past stem + inflectional suffix (+ derivational suffix)
- derivational prefix + past stem + inflectional suffix (+ derivational suffix)
- derivational prefix + inflectional prefix + past stem + inflectional suffix (+ derivational suffix)

In addition, a few examples are provided for each rule. The morphological rules aim at analyzing the underlying representation of each word-form. Following this step, morphophonological rules are applied to the surface representation. These rules are described in Section 6 and specified, whenever used within the rules, by ⇒. Therefore, allomorphs which are created using morphophonological rules are not included in these morphological rules.

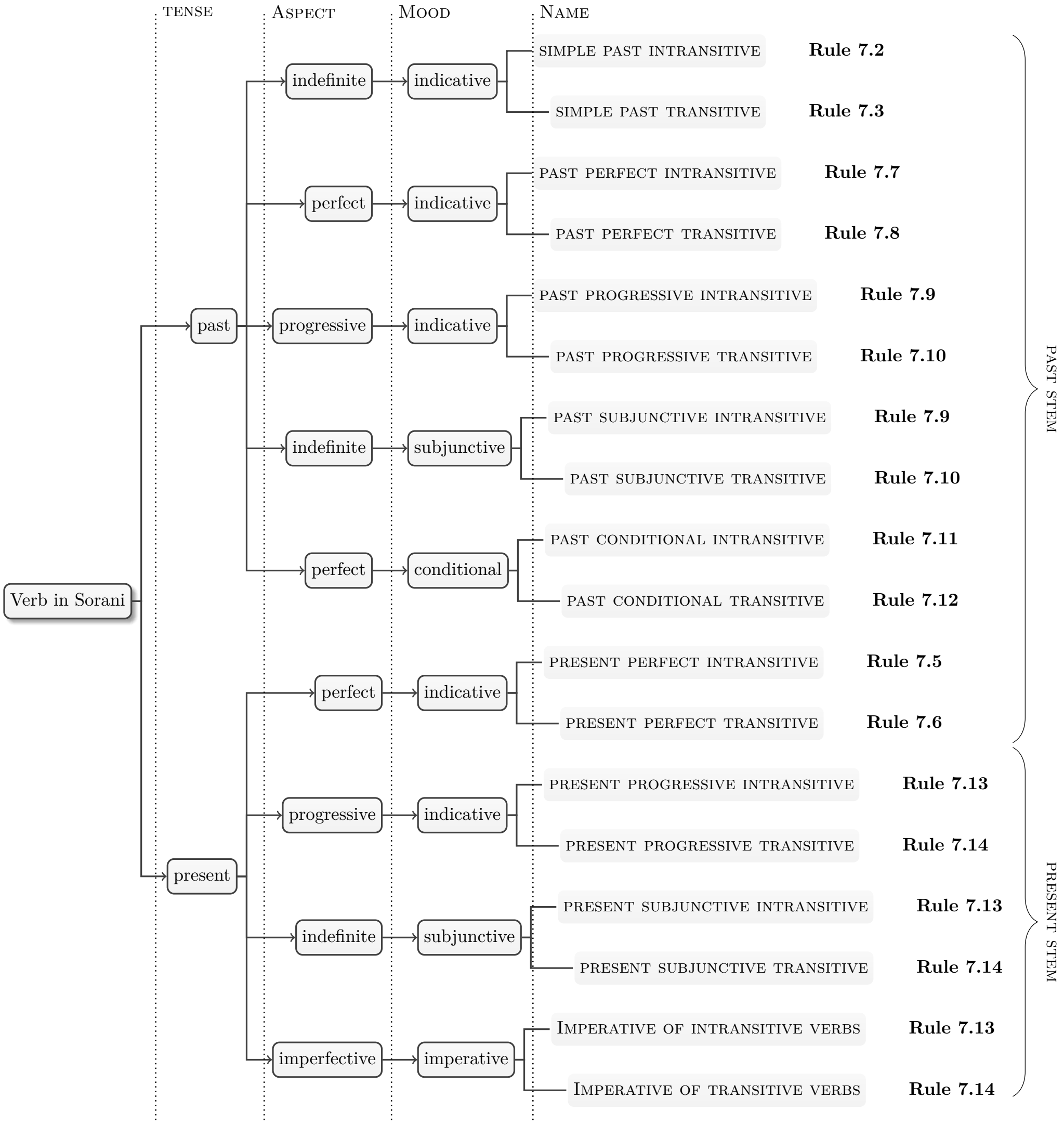

Figure 4: Tenses in Sorani Kurdish according to their constructing stems from an inflectional point of view

## 7.1 Verbs

Verbs in Sorani Kurdish can represent tense, aspect and mood. The majority of verb lexemes are created using derivational affixes, such as verbal particles and absolute adpositions. Therefore, verb forms can get easily complex, particularly due to the erratic patterns of pronominal endoclitics and emphasis endoclitic, i.e. *=îş*.

Figure 4 depicts the possible verbal constructions of Sorani Kurdish. For each construction a rule is defined. In order to keep the rules concise, we use PG and SG which denote a prefix group and a suffix group, respectively. Such groups may be a compound of inflectional and derivational morphemes which are specified for each rule independently. In addition, we use the following acronyms to avoid redundancy in the rules. These acronyms are previously described in Section 5.

| Abbreviation | Paradigm | Description | |
|---|---|---|---|
| | | Type | Value |
| $DP_1$ | derivational | verbal particles (Table 11) | *ber-, da-, der-, hel-, ser-, řa-, řo-, wer-* |
| | derivational | absolute adposition (Table 14) | *bo, lê, pê, tê, wê, legeł, degeł, letek* |
| | derivational | absolute reciprocal adposition (Table 15) | *lêk, pêk, têk, wêk* |
| $DS_{PST}$ | derivational | past stem suffixes (Section 5.1.4) | *∅, -a, -and, -ard, -d, -î, -îşt, -ird, -t, -û, -y* |
| $DS_{PRS}$ | derivational | present stem suffixes (Section 5.1.4) | *∅, -ê, -ên, -êr* |

Table 16: Description of some of the abbreviations used in the rules with references

**Rule 7.1.** Infinitive of transitive and intransitive verbs

| | | |
|---|---|---|
| (a) word-form | = $PG_1$ + past stem + inflectional suffix (+ derivational suffix) | |
| (b) $PG_1$ | = (derivational prefix +) (+ inflectional prefix +) | |
| (c) derivational prefix | = $DP_1$ | (Table 16) |
| (d) inflectional prefix | = *ne-* | NEGATION PREFIX |
| (e) past stem | = root + derivational suffix | |
| (f) root | = *awis, bi, çi, ge, haw, kew, kêł, ki, sût, şar, ...* | |
| (g) derivational suffix | = $DS_{PST}$ | (Table 16) |
| (h) inflectional suffix | = *-in* | |
| (i) derivational suffix | = *-ewe* | REPETITION SUFFIX |
| | → *hên-a-in* ⇒ *hênan* (to bring) | Rule 6.1(b) |
| - word-form: | → *ne-hat-in* ⇒ *nehatin* (not to come) | |
| | → *xw-ard-in-ewe* ⇒ *xwardinewe* (to eat again; to drink) | |

**Rule 7.2.** Simple past (or preterite) of intransitive verbs

| | | |
|---|---|---|
| (a) word-form | = $PG_1$ + past stem + $SG_1$ | |
| (b) $PG_1$ | = (derivational prefix +) (+ inflectional prefix +) | |
| (c) derivational prefix | = $DP_1$ | (Table 16) |
| (d) inflectional prefix | = *ne-* | NEGATION PREFIX |
| (e) past stem | = root + derivational suffix | |
| (f) root | = *awis, bi, çi, ge, kew...* | |
| (g) derivational suffix | = $DS_{PST}$ | (Table 16) |
| (h) $SG_1$ | = (i) inflectional suffix (+inflectional suffix) (+repetition suffix)<br>= (ii) inflectional suffix (+inflectional suffix) (+ derivational suffix) | |
| (i) inflectional suffix | = *-im, -î/ît, ∅, -în, -in* | AGENT SUFFIX |
| (j) inflectional suffix | = *-e* | DIRECTIONAL SUFFIX |
| (k) inflectional suffix | = *-ewe* | REPETITION SUFFIX |
| (l) derivational suffix | = *-ewe* | REPETITION SUFFIX |
| | → *ser-kewt-im* ⇒ *serkewtim* ((I) succeeded) | |
| - word-form: | → *ge-îşt-în-ewe* ⇒ *geyîştînewe* ((we) arrived back) | Rule 6.1(c) |
| | → *çû-∅-e* ⇒ *çûwe* ((he/she) went to) | Rule 6.1(c) |

**Rule 7.3.** Simple past of transitive verbs

| | | |
|---|---|---|
| (a) word-form | = (i) past stem + inflectional suffix + $SG_2$<br>= (ii) $PG_1$ + past stem + $SG_2$ | |
| (b) past stem | = root + derivational suffix | |
| (c) root | = *bi, haw, hên, sût...* | |
| (d) derivational suffix | = $DS_{PST}$ | (Table 16) |
| (e) inflectional suffix | = *-im, -it, î, -man, -tan, -yan* | AGENT SUFFIX |
| (f) $SG_2$ | = inflectional suffix + $SG_3$ | |
| (g) inflectional suffix | = *-im, -î/ît, ∅, -în, -in* | PATIENT SUFFIX |
| (f) $SG_3$ | = (i) (+ inflectional suffix +) (+inflectional suffix)<br>= (ii) (+ inflectional suffix +) (+ derivational suffix) | |
| (h) inflectional suffix | = *-e* | DIRECTIONAL SUFFIX |
| (i) inflectional suffix | = *-ewe* | REPETITION SUFFIX |
| (j) derivational suffix | = *-ewe* | REPETITION SUFFIX |
| (k) $PG_1$ | = derivational prefix + inflectional prefix (+ inflectional prefix) | |
| (l) derivational prefix | = $DP_1$ | (Table 16) |
| (m) inflectional prefix | = *im-, it-, î-, man-, tan-, yan-* | AGENT PREFIX |
| (n) inflectional prefix | = *ne-* | NEGATION PREFIX |
| | → *hên-a-im-e* ⇒ *hêname* ((I) brought to) | Rule 6.1(b) |
| - word-form: | → *ne-yan-bi-ird-în-ewe* ⇒ *neyanbirdînewe* ((they) did not bring us again/back) | |
| | → *da-î-sût-and-im* ⇒ *daysûtandim* ((he, she) burnt me down) Rule 6.1(c) | |

**Rule 7.4.** Perfect active participle

| | | |
|---|---|---|
| (a) word-form | = $PG_1$ + past stem + inflectional suffix | |
| (b) $PG_1$ | = (derivational prefix +) (+ inflectional prefix +) | |
| (c) derivational prefix | = $DP_1$ | (Table 16) |
| (e) inflectional prefix | = *ne-* | NEGATION PREFIX |
| (f) past stem | = root + derivational suffix | |
| (g) root | = *awis, bi, çi, ge, kew...* | |
| (h) derivational suffix | = $DS_{PST}$ | (Table 16) |
| (i) inflectional suffix | = *-û* | |
| | → *ge-îşt-û* ⇒ *geyîştû* (arrived) | Rule 6.1(c) |
| - word-form: | → *da-ma-û* ⇒ *damaw* (remained down; miserable (adj.)) | Rule 6.1(c) |
| | → *bi-ird-û* ⇒ *birdû* (taken) | Rule 6.1(a) |

**Rule 7.5.** Present perfect of intransitive verbs

| | | |
|---|---|---|
| (a) word-form | = $PG_1$ + past stem + inflectional suffix + $SG_1$ | |
| (b) $PG_1$ | = (derivational prefix +) (+ inflectional prefix +) | |
| (c) derivational prefix | = $DP_1$ | (Table 16) |
| (d) inflectional prefix | = *ne-* | NEGATION PREFIX |
| (e) past stem | = root + derivational suffix | |
| (f) root | = *awis, bi, çi, ge, kew...* | |
| (g) derivational suffix | = $DS_{PST}$ | (Table 16) |
| (h) inflectional suffix | = *-û* | |
| (i) $SG_1$ | = (i) inflectional suffix (+ inflectional suffix +) (+inflectional suffix)<br>= (ii) inflectional suffix (+ inflectional suffix +) (+ derivational suffix) | |
| (j) inflectional suffix | = *-im, -î/ît, -e, -în, -in* | AGENT SUFFIX |
| (k) inflectional suffix | = *-e* | DIRECTIONAL SUFFIX |
| (l) inflectional suffix | = *-ewe* | REPETITION SUFFIX |
| (m) derivational suffix | = *-ewe* | REPETITION SUFFIX |
| - word-form: | → *ge-îşt-û-im* ⇒ *geyîştûm* ((I) have arrived) Rule 6.1(b & c)<br>→ *heł-hat-û-e-ewe* ⇒ *hełhatûwetewe* (has fled away to) Rule 6.1(c) & Rule 6.2(a)<br>→ *ne-ma-û-î-e* ⇒ *nemawîye* (have.2SG not remained from/to) Rule 6.1(c) | |

**Rule 7.6.** Present perfect of transitive verbs

| | | |
|---|---|---|
| (a) word-form | = (i) past stem + $SG_1$ + $SG_2$ + $SG_3$<br>= (ii) $PG_1$ + past stem + $SG_2$ + $SG_3$ | |
| (b) past stem | = root + derivational suffix | |
| (c) root | = *bi, haw, hên, sût...* | |
| (d) derivational suffix | = $DS_{PST}$ | (Table 16) |
| (e) $SG_1$ | = inflectional suffix + inflectional suffix | |
| (f) inflectional suffix | = *-û* | |
| (g) inflectional suffix | = *-im, -it, î, -man, -tan, -yan* | AGENT SUFFIX |
| (h) $SG_2$ | = inflectional suffix + inflectional suffix | |
| (i) inflectional suffix | = *-im, -î/ît, ∅, -în, -in* | PATIENT SUFFIX |
| (j) inflectional suffix | = *e* | 3SG COPULA |
| (k) $SG_3$ | = (i) (+ inflectional suffix +) (+inflectional suffix)<br>= (ii) (+ inflectional suffix +) (+ derivational suffix) | |
| (l) inflectional suffix | = *-e* | DIRECTIONAL SUFFIX |
| (m) inflectional suffix | = *-ewe* | REPETITION SUFFIX |
| (n) derivational suffix | = *-ewe* | REPETITION SUFFIX |
| (o) $PG_1$ | = derivational prefix + inflectional prefix (+ inflectional prefix) | |
| (p) derivational prefix | = $DP_1$ | (Table 16) |
| (q) inflectional prefix | = *im-, it-, î-, man-, tan-, yan-* | AGENT PREFIX |
| (r) inflectional prefix | = *ne-* | NEGATION PREFIX |
| - word-form: | → *hên-a-û-im-e-e* ⇒ *hênawmete* ((I) have brought to) Rule 6.1(c) & 6.2(a)<br>→ *ne-man-sût-and-û-e* ⇒ *nemansûtandûwe* ((we) have not burnt it) Rule 6.1(c)<br>→ *bi-ird-û-im-in-e* ⇒ *birdûmine* ((I) have taken them) Rule 6.1(a & b) | |

**Rule 7.7.** Past perfect (indicative) and Past subjunctive of intransitive verbs

| | | |
|---|---|---|
| (a) word-form | = $PG_1$ + past stem + $SG_1$ + $SG_2$ | |
| (b) $PG_1$ | = (derivational prefix +) (+ inflectional prefix +) | |
| (c) derivational prefix | = $DP_1$ | (Table 16) |
| (d) inflectional prefix | = *ne-* | NEGATION PREFIX |
| (e) past stem | = root + derivational suffix | |
| (f) root | = *awis, bi, çi, ge, kew...* | |
| (g) derivational suffix | = $DS_{PST}$ | (Table 16) |
| () $SG_1$ | inflectional suffix + inflectional suffix + inflectional suffix | |
| (h) inflectional suffix | = *-i* | |
| (i) inflectional suffix | = (i) *bû* = 3SG.PAST.BÛN (to be)<br>= (ii) *bêt* = 3SG.PRESENT.BÛN (to be) | INDICATIVE<br>SUBJUNCTIVE |
| (k) inflectional suffix | = *-im, -î/ît, -e, -în, -in* | AGENT SUFFIX |
| (j) $SG_2$ | = (i) (+ inflectional suffix +) (+inflectional suffix)<br>= (ii) (+ inflectional suffix +) (+ derivational suffix) | |
| (l) inflectional suffix | = *-e* | DIRECTIONAL SUFFIX |
| (m) inflectional suffix | = *-ewe* | REPETITION SUFFIX |
| (n) derivational suffix | = *-ewe* | REPETITION SUFFIX |
| - word-form: | → *ge-îşt-i-bû-im* ⇒ *geyîştibûm* ((I) had arrived) Rule 6.1(c & b)<br>→ *heł-hat-i-bû* ⇒ *hełhatibû* ((he, she) had fled away )<br>→ *ne-ma-i-bû-în-e* ⇒ *nemabûyne* ((you.2SG) had not remained from/to) Rule 6.1(b & c) | |

**Rule 7.8.** Past perfect (indicative) and past subjunctive of transitive verbs

| | | |
|---|---|---|
| (a) word-form | = (i) past stem + $SG_1$ + inflectional suffix + $SG_2$<br>= (ii) $PG_1$ + past stem + $SG_1$ + $SG_2$ | |
| (b) past stem | = root + derivational suffix | |
| (c) root | = *bi, haw, hên, sût...* | |
| (d) derivational suffix | = $DS_{PST}$ | (Table 16) |
| (e) $SG_1$ | = inflectional suffix + inflectional suffix | |
| (f) inflectional suffix | = *-i* | |
| (g) inflectional suffix | = (i) *bû* = 3SG.PAST.BÛN (to be)<br>= (ii) *bêt* = 3SG.PRESENT.BÛN (to be) | INDICATIVE<br>SUBJUNCTIVE |
| (h) inflectional suffix | = *-im, -it, î, -man, -tan, -yan* | AGENT SUFFIX |
| (i) $SG_2$ | = inflectional suffix + $SG_3$ | |
| (k) inflectional suffix | = *-im, -î/ît, ∅, -în, -in* | PATIENT SUFFIX |
| (l) $SG_3$ | = (i) (+ inflectional suffix +) (+inflectional suffix)<br>= (ii) (+ inflectional suffix +) (+ derivational suffix) | |
| (m) inflectional suffix | = *-e* | DIRECTIONAL SUFFIX |
| (n) inflectional suffix | = *-ewe* | REPETITION SUFFIX |
| (o) derivational suffix | = *-ewe* | REPETITION SUFFIX |
| (p) $PG_1$ | = derivational prefix + inflectional prefix (+ inflectional prefix) | |
| (q) derivational prefix | = $DP_1$ | (Table 16) |
| (r) inflectional prefix | = *im-, it-, î-, man-, tan-, yan-* | AGENT PREFIX |
| (s) inflectional prefix | = *ne-* | NEGATION PREFIX |
| - word-form: | → *hên-a-i-bû-im-e* ⇒ *hênabûme* ((I) had brought to)<br>→ *ne-man-sût-and-i-bû* ⇒ *nemansûtandibû* ((we) had not burnt it)<br>→ *bi-ird-û-im-in* ⇒ *birdûmine* ((I) have taken them) | Rule 6.1(b)<br><br>Rule 6.1(a) |

**Rule 7.9.** Past progressive of intransitive verbs

| | | |
|---|---|---|
| (a) word-form | = $PG_1$ + present stem + $SG_1$ | |
| (b) $PG_1$ | = (derivational prefix +) (+ inflectional prefix +) inflectional prefix | |
| (c) derivational prefix | = $DP_1$ | (Table 16) |
| (d) inflectional prefix | = *ne-* | NEGATION PREFIX |

| | |
|---|---|
| (d) inflectional prefix | = *de-,e-* MODAL PREFIX |
| (e) present stem | = root + derivational suffix |
| (f) root | = *awis, bi, çi, ge, haw, kew, kêł, ki, sût, şar, ...* |
| (g) derivational suffix | = $DS_{PRS}$ (Table 16) |
| (h) $SG_1$ | = (i) inflectional suffix (+ inflectional suffix +) (+inflectional suffix)<br>= (ii) inflectional suffix (+ inflectional suffix +) (+ derivational suffix) |
| (i) inflectional suffix | = *-im, -î/ît, ∅, -în, -in* AGENT SUFFIX |
| (j) inflectional suffix | = *-e* DIRECTIONAL SUFFIX |
| (k) inflectional suffix | = *-ewe* REPETITION SUFFIX |
| (l) derivational suffix | = *-ewe* REPETITION SUFFIX |
| - word-form: | → *de-çi-û-im* ⇒ *deçûm* ((I) was going) Rule 6.1(b)<br>→ *e-ma-∅-ewe* ⇒ *emayewe* ((he, she) was staying) Rule 6.1(c)<br>→ *hel-ne-de-hat-în-ewe* ⇒ *helnedehatînewe* ((we) were not fleeing away) |

**Rule 7.10.** Past progressive of transitive verbs

| | |
|---|---|
| (a) word-form | = $PG_1$ + past stem + inflectional suffix + $SG_1$ |
| (b) $PG_1$ | = (i) inflectional $prefix_1$ + inflectional $prefix_2$<br>= (ii) inflectional $prefix_3$ + inflectional $prefix_2$ + inflectional $prefix_1$<br>= (iii) derivational prefix + inflectional $prefix_2$ (+ inflectional $prefix_3$) + inflectional $suffix_1$ |
| (c) inflectional $prefix_1$ | = *de-,e-* MODAL PREFIX |
| (d) inflectional $prefix_2$ | = *im-, it-, î-, man-, tan-, yan-* AGENT PREFIX |
| (e) inflectional $prefix_3$ | = *ne-* NEGATION PREFIX |
| (h) derivational prefix | = $DP_1$ (Table 16) |
| (i) past stem | = root + derivational suffix |
| (j) root | = *bi, haw, hên, sût...* |
| (d) derivational suffix | = $DS_{PST}$ (Table 16) |
| (g) inflectional suffix | = *-im, -î/ît, ∅, -în, -in* PATIENT SUFFIX |
| (f) $SG_1$ | = (i) (+ inflectional suffix +) (+inflectional suffix)<br>= (ii) (+ inflectional suffix +) (+ derivational suffix) |
| (h) inflectional suffix | = *-e* DIRECTIONAL SUFFIX |
| (i) inflectional suffix | = *-ewe* REPETITION SUFFIX |
| (j) derivational suffix | = *-ewe* REPETITION SUFFIX |
| - word-form: | → *de-im-hên-a-e* ⇒ *demhênaye* ((I) was bringing to) Rule 6.2(b & c)<br>→ *ne-yan-de-kêł-a* ⇒ *neyandekêła* ((they) were not ploughing)<br>→ *da-î-de-sût-and-im* ⇒ *daydesûtandim* ((he, she) burnt me down) Rule 6.1(c) |

**Rule 7.11.** Past conditional of intransitive verbs

| | |
|---|---|
| (a) word-form | = $PG_1$ + past stem + $SG_1$ + inflectional $suffix_1$ + $SG_2$<br>= $PG_1$ + past stem + $SG_3$ + $SG_2$ |
| (b) $PG_1$ | = (derivational prefix +) (+ inflectional prefix +) |
| (c) derivational prefix | = $DP_1$ (Table 16) |
| (d) inflectional prefix | = *bi-, ne-* |
| (e) past stem | = root + derivational suffix |
| (f) root | = *awis, bi, çi, ge, kew...* |
| (g) derivational suffix | = $DS_{PST}$ (Table 16) |
| (h) $SG_1$ | = inflectional suffix + inflectional suffix |

| | | |
|---|---|---|
| (i) inflectional suffix | = *-i* | |
| (j) inflectional suffix | = *ba* | 3SG.PAST.SUBJ.BÛN (to be) |
| (k) inflectional suffix$_1$ | = *-im, -î/ît, -e, -în, -in* | AGENT SUFFIX |
| (l) SG$_2$ | = (i) (+ inflectional suffix +) (+inflectional suffix)<br>= (ii) (+ inflectional suffix +) (+ derivational suffix) | |
| (m) inflectional suffix | = *-e* | DIRECTIONAL SUFFIX |
| (n) inflectional suffix | = *-ewe* | REPETITION SUFFIX |
| (o) derivational suffix | = *-ewe* | REPETITION SUFFIX |
| (p) SG$_3$ | = inflectional suffix$_1$ + inflectional suffix | |
| (q) inflectional suffix | = *-aye* | |
| - word-form: | → *ge-îşt-i-ba-im* ⇒ *geyîştibûm* ((I) would had arrived) Rule 6.1(b & c)<br>→ *hel-hat-i-în-aye* ⇒ *hełhatînaye* ((we) would had fled away) Rule 6.1(b)<br>→ *ne-ma-i-ba-în-e* ⇒ *nemabayne* ((you.2SG) have not remained from/to) Rule 6.1(b & c) | |

**Rule 7.12.** Past conditional of transitive verbs

| | | |
|---|---|---|
| (a) word-form | = (i) past stem + SG$_1$ + inflectional suffix + SG$_2$<br>= (ii) PG$_1$ + past stem + SG$_1$ + SG$_2$ | |
| (b) past stem | = root + derivational suffix | |
| (c) root | = *bi, haw, hên, sût...* | |
| (d) derivational suffix | = DS$_{PST}$ | (Table 16) |
| (e) SG$_1$ | = inflectional suffix + inflectional suffix | |
| (f) inflectional suffix | = *-i* | |
| (g) inflectional suffix | = *ba* | 3SG.PAST.SUBJ.BÛN (to be) |
| (h) inflectional suffix | = *-im, -it, î, -man, -tan, -yan* | AGENT SUFFIX |
| (i) SG$_2$ | = inflectional suffix + SG$_3$ | |
| (k) inflectional suffix | = *-im, -î/ît, ∅, -în, -in* | PATIENT SUFFIX |
| (l) SG$_3$ | = (i) (+ inflectional suffix +) (+inflectional suffix)<br>= (ii) (+ inflectional suffix +) (+ derivational suffix) | |
| (m) inflectional suffix | = *-e* | DIRECTIONAL SUFFIX |
| (n) inflectional suffix | = *-ewe* | REPETITION SUFFIX |
| (o) derivational suffix | = *-ewe* | REPETITION SUFFIX |
| (p) PG$_1$ | = (i) derivational prefix + inflectional prefix$_1$ (+ inflectional prefix$_2$)<br>= (ii) inflectional prefix$_2$ + inflectional prefix$_1$ | |
| (q) derivational prefix | = DP$_1$ | (Table 16) |
| (r) inflectional prefix$_1$ | = *im-, it-, î-, man-, tan-, yan-* | AGENT PREFIX |
| (s) inflectional prefix$_2$ | = *bi-, ne-* | |
| - word-form: | → *hên-a-i-ba-im-e-e* ⇒ *hênabame* ((I) would had brought to) Rule 6.1(b)<br>→ *bi-im-bi-ird-i-ba-in* ⇒ *bimbirdiba* ((I) would have taken) Rule 6.1(a & b)<br>→ *ne-man-sût-and-i-ba* ⇒ *nemansûtandiba* ((we) would had not burnt it) | |

**Rule 7.13.** Present progressive (indicative) and present subjunctive and imperative of intransitive verbs[8]

| | |
|---|---|
| (a) word-form | = PG$_1$ + present stem + inflectional suffix + SG$_1$ |

[8] HATIN (to come) and ÇÛN (to go) have exceptional imperative forms as *were*/*bê* and *biçô*, respectively

| | | |
|---|---|---|
| (b) $PG_1$ | = (derivational prefix +) inflectional prefix | |
| (c) derivational prefix | = $DP_1$ | (Table 16) |
| (d) inflectional prefix | = (i) *de-,e-, na-* | INDICATIVE |
| | = (ii) *bi-, ne-, ∅* | SUBJUNCTIVE |
| | = (iii) *bi-, me-* | IMPERATIVE |
| (e) present stem | = root + derivational suffix | |
| (f) root | = *awis, bi, çi, ge, haw, kew, kêl, ki, sût, şar, ...* | |
| (g) derivational suffix | = $DS_{PRS}$ | (Table 16) |
| (h) inflectional suffix | = (i) *-im, -î/ît, -ê/êt, -în, -in* | AGENT SUFFIX (NOT IMP) |
| | = (ii) *-e, -in* | AGENT SUFFIX (ONLY IMP) |
| (i) $SG_1$ | = (i) (+ inflectional suffix +) (+inflectional suffix) | |
| | = (ii) (+ inflectional suffix +) (+ derivational suffix) | |
| (j) inflectional suffix | = *-e* | DIRECTIONAL SUFFIX |
| (k) inflectional suffix | = *-ewe* | REPETITION SUFFIX |
| (l) derivational suffix | = *-ewe* | REPETITION SUFFIX |
| - word-form: | → *e-çi-în-e* ⇒ *eçîne* ((we) go to) | Rule 6.1(b) |
| | → *de-ge-ê* ⇒ *dega* ((he/she) arrives) | Rule 6.2(c) |
| | → *bi-çi-în-ewe* ⇒ *biçînewe* ((that we) go) | Rule 6.1(b) |
| | → *bi-ge-in* ⇒ *bigene* ((you.PL) arrive!) | Rule 6.2(b) |
| | → *me-kew-e* ⇒ *mekewe* ((you.SG) don't fall!) | |

**Rule 7.14.** Present progressive (indicative) and present subjunctive and imperative of transitive verbs[9]

| | | |
|---|---|---|
| (a) word-form | = $PG_1$ + present stem + inflectional suffix + $SG_1$ | |
| (b) $PG_1$ | = (i) inflectional prefix$_1$ ( + inflectional prefix$_2$ ) | |
| | = (ii) derivational prefix ( + inflectional prefix$_2$ + ) + inflectional prefix$_1$ | |
| (c) inflectional prefix$_1$ | = (i) *de-,e-, na-* | INDICATIVE |
| | = (ii) *bi-, ne-, ∅* | SUBJUNCTIVE |
| | = (iii) *bi-, me-* | IMPERATIVE |
| (d) inflectional prefix$_2$ | = *-im, -it, -î, -man, -tan, -yan* | PATIENT PREFIXES |
| (e) derivational prefix | = $DP_1$ | (Table 16) |
| (f) present stem | = root + derivational suffix | |
| (g) root | = *awis, bi, çi, ge, haw, kew, kêl, ki, sût, şar, ...* | |
| (h) derivational suffix | = $DS_{PRS}$ | (Table 16) |
| (i) inflectional suffix | = (i) *-im, -î/ît, -ê/êt, -în, -in* | AGENT SUFFIX (NOT IMP) |
| | = (ii) *-e, -in* | AGENT SUFFIX (ONLY IMP) |
| (j) $SG_1$ | = (i) (+ inflectional suffix +) (+ inflectional suffix ) | |
| | = (ii) (+ inflectional suffix +) (+ derivational suffix) | |
| (k) inflectional suffix | = *-e* | DIRECTIONAL SUFFIX |
| (l) inflectional suffix | = *-ewe* | REPETITION SUFFIX |
| (m) derivational suffix | = *-ewe* | REPETITION SUFFIX |
| - word-form: | → *e-sût-ên-im* ⇒ *esûtênim* ((I) burn/ I am burning) | |
| | → *da-î-de-hên-in* ⇒ *daydehênin* ((they) invent it) | Rule 6.1(c) |
| | → *bi-yan-be-î* ⇒ *biyanbey* ((that you.2SG) take them) | Rule 6.1(c) |
| | → *na-xo-êt-ewe* ⇒ *naxwatewe* ((he, she) is not drinking) | Rule 6.2(d) |
| | → *me-xo-in-ewe* ⇒ *mexonewe* ((you.PL) don't drink!) | Rule 6.1(b) |
| | → *bi-î-ke-e-ewe* ⇒ *bîkerewe* ((you.SG) open!) | Rule 6.1(b) & Rule 6.2 (b) |

[9]Subjunctive forms are used as imperative for passive verbs

# 8 Conclusion and Future Work

In this paper, we describe the Sorani Kurdish morphology from a computational point of view. First, we describe morphemes in Sorani Kurdish as bound morphemes and free morphemes. In the first type of morphemes, we cover clitics, such as pronominal endoclitics, endoclitic of emphasis *=îş*, postposed directional complement *=e* and pronominal adverb *=ê*, and affixes, such as inflectional and derivational prefixes and suffixes. Regarding free morphemes, we present adpositions with their simple and absolute forms, and also discuss how pronouns vary among Sorani sub-dialects. Furthermore, we present morphological and morphophonological rules which can be used as finite-state transducers for the morphological analysis of Sorani Kurdish. We cover four open-class part-of-speech categories, namely verbs, nouns, adjectives, adverbs and prepositions.

Given the complexity of Sorani Kurdish morphology, the current study can be used as a comprehensive reference which can be implemented for morphological analysis. Morphological analysis is an important component of various NLP applications, particularly lexical and syntactic analysis, and can pave the way for further progress in language technology for Sorani Kurdish.

As future work, creating a morphological analyzer based on the defined rules is proposed. Having various sub-dialects, Sorani Kurdish represents a diverse range of features in morphology and syntax. For instance, the placement of the pronominal endoclitics and endoclitic of emphasis *=îş* varies depending on the sub-dialect. Describing this diversity should also be the addressed in a future work. In the same vein, we believe that other dialects of Kurdish, i.e. Kurmanji and Southern Kurdish, along with Zazaki and Gorani language should be addressed from a computational morphology perspective.

# 9 Acknowledgements

The author would like to thank Dr. Francis M. Tyers and Dr. Theodorus Fransen for their constructive comments and feedback on this manuscript.

# Appendices

## A List of Bound Morphemes in Sorani Kurdish

∅ absolute singular and plural, nominative, 3sg pronominal enclitic, past and present stem derivational suffix
**-a/ا** past stem derivational suffix
**-an/ان** plural indefinite
**-and/اند** past stem derivational suffix
**-ane/انە** demonstrative plural
**-ard/ارد** past stem derivational suffix
**-at/ات** plural indefinite
**-d/د** past stem derivational suffix
**-diran/دران** past passive voice derivational suffix
**-dirê/درێ** present passive voice derivational suffix
**-e/ە** demonstrative singular, masculine vocative, allomorph of Izafa, present stem derivational suffix
**-ekan/ەکان** plural definite
**-eke/ەکە** singular definite
**-ewe/ەوە** postposition, inflectional and derivational verbal suffix
**-gel/گەل** plural indefinite
**-gele/گەلە** demonstrative plural
**-ha/ها** plural indefinite
**-i** auxiliary morpheme in perfective tenses
**-in/ن** infinitive marker
**-ird/رد** past stem derivational suffix
**-ran/ران** allomorph of *-diran*
**-rewe/رەوە** allomorph of *-ewe*
**-rê/رێ** allomorph of *-dirê*
**-t/ت** past stem derivational suffix
**-tir/تر** comparative
**-tirîn/ترین** superlative
**-y/ی** past stem derivational suffix
**-yek/یەک** allomorph of *-ek*
**-ê/ێ** feminine oblique, feminine vocative, locative, present stem derivational suffix
**-êk/ێک** singular indefinite
**-ên/ێن** present stem derivational suffix
**-êr/ێر** present stem derivational suffix
**-î/ی** masculine oblique, Izafa, past stem derivational suffix
**-îne/ینە** plural vocative
**-îşt/یشت** past stem derivational suffix
**-û/وو** past stem derivational suffix, auxiliary morpheme in past participle verbs
**=a/ا** allomorph of *=ê* pronominal enclitic
**=at/ات** allomorph of *=êt* pronominal enclitic
**=e/ە** 3sg copula, 2sg imperative pronominal enclitic, postposed directional complement
**=im/م** 1sg copula, 1sg pronominal enclitic and endoclitic
**=in/ن** 2pl & 3pl copula, 2pl & 3pl pronominal enclitic, 2pl imperative pronominal enclitic
**=it/ت** 2sg pronominal endoclitic
**=man/مان** 1pl pronominal endoclitic
**=tan/تان** 2pl pronominal endoclitic
**=te/تە** allomorph of *=e* postposed directional complement
**=wa/وا** allomorph of *=ê* pronominal enclitic
**=wat/وات** allomorph of *=êt* pronominal enclitic
**=yan/یان** 3pl pronominal endoclitic
**=ê/ێ** 3sg pronominal enclitic, pronominal adverb
**=êt/ێت** 3sg pronominal enclitic
**=î/ی** 2sg copula, 2sg pronominal enclitic, 3sg pronominal endoclitic
**=în/ین** 1pl copula, 1sg pronominal enclitic
**=ît/یت** 2sg copula, 2sg pronominal enclitic
**=îş/یش** emphasis endoclitic
**=ş/ش** allomorph of *=îş*

**ber-/بەر** derivational verbal particle
**bi-/ب** subjunctive and imperative marker

**da-/دا** derivational verbal particle
**de-/دە** progressive marker
**der-/دەر** derivational verbal particle

**e-/ئە** progressive marker

**heł-/هەڵ** derivational verbal particle

**me-/مە** negative imperative marker

**na-/نا** negative marker
**ne-/نە** negative marker

**ser-/سەر** derivational verbal particle

**we-/وە** subjunctive marker
**wer-/وەر** derivational verbal particle

**řa-/ڕا** derivational verbal particle
**řo-/ڕۆ** derivational verbal particle

# B Sample Paradigms

In the following, a few word-forms are analyzed according to the rules described in Section 6 and Section 7.

**Example B.1.** Construction of *kewtin* (FALL.INF)

- word-form → stem + inflectional suffix — Rule 7.1(a)
- stem → root + derivational suffix — Rule 7.1(e)
- root → *kew* — Rule 7.1(f)
- derivational suffix → *-t* — Rule 7.1(g)
- inflectional suffix → *-in* — Rule 7.1(h)
- word-form: *kew-t-in* ⇒ *kewtin*

**Example B.2.** Construction of *hełkewt* (HAPPEN.PST.3SG)

- word-form → derivational prefix + stem + inflectional suffix — Rule 7.2(a) & (b)
- derivational prefix → *heł-* — Rule 7.2(c)
- stem → root + derivational suffix — Rule 7.2(e)
- root → *kew* — Rule 7.2(f)
- derivational suffix → *-t* — Rule 7.2(g)
- inflectional suffix → ∅ — Rule 7.2(i)
- word-form: *heł-kew-t-in* ⇒ *hełkewtin*

**Example B.3.** Construction of *bîyanbîninewe* (FIND.SBJV.1PL)

- word-form → $PG_1$ + present stem + inflectional suffix + $SG_1$ — Rule 7.14(a)
- $PG_1$ → inflectional $prefix_1$ + inflectional $prefix_2$ — Rule 7.14(b/i)
- inflectional $prefix_1$ → *bi-* — Rule 7.14(c/ii)
- inflectional $prefix_2$ → *=yan* — Rule 7.14(d)
- present stem → root + derivational suffix — Rule 7.14(f)
- root → *bîn* — Rule 7.14(g)
- derivational suffix → ∅ — Rule 7.14(h)
- inflectional suffix → *=în* — Rule 7.14(i/i)
- $SG_1$ → derivational suffix — Rule 7.14(j/ii)
- derivational suffix → *-ewe* — Rule 7.14(m)
- word-form: *bi-yan-bîn-în-ewe* ⇒ *bî-yan-bîn-în-ewe* — Rule 6.1(c)

**Example B.4.** Construction of *bige* (ARRIVE.IMP.2SG)

- word-form → $PG_1$ + present stem + inflectional suffix — Rule 7.13(a)
- $PG_1$ → inflectional prefix — Rule 7.13(b)
- inflectional $prefix_1$ → *bi-* — Rule 7.13(D/ii)
- present stem → root + derivational suffix — Rule 7.13(e)
- root → *ge* — Rule 7.13(f)
- derivational suffix → ∅ — Rule 7.13(g)
- inflectional suffix → *=e* — Rule 7.13(h/ii)
- word-form: *bi-ge* ⇒ *bige*